\documentclass[runningheads]{llncs}

\usepackage[mobile]{eccv}

\usepackage{eccvabbrv}

\usepackage{graphicx}
\usepackage{booktabs}
\usepackage{multirow}
\usepackage[table]{xcolor}
\usepackage{soul}
\usepackage{booktabs}    % For professional horizontal rules
\usepackage{tabularx}    % For flexible column widths
\usepackage{amsmath}     % For math symbols
\usepackage{caption}     % For table captions
\usepackage{colortbl}    % For row/cell coloring if needed
\usepackage[normalem]{ulem} % For underlining
\usepackage{makecell}
\usepackage{enumitem}
\usepackage[accsupp]{axessibility}  % Improves PDF readability for those with disabilities.
\usepackage{enumitem}

\usepackage[pagebackref,breaklinks,colorlinks,citecolor=eccvblue]{hyperref}
\usepackage{orcidlink}

\begin{document}

\title{RealVDeblur: One-Step Diffusion for Generalizable Real-World Video Deblurring} 

\titlerunning{RealVDeblur}

\author{Renbiao Jin\inst{1,2,*} \and
Mingxin Yang\inst{1,2,*} \and Yutian Chen\inst{2,3} \and \\ Junhao Zhuang\inst{5} \and  Xin Cai\inst{2,3} \and Mulin Yu\inst{2} \and Linning Xu\inst{3} \and \\ Wenxian Yu\inst{1} \and  Danping Zou\inst{1,\dag} \and Shi Guo\inst{2,\dag} \and Tianfan Xue\inst{3,2,4}}

\renewcommand{\thefootnote}{\fnsymbol{footnote}}
\footnotetext[0]{\hspace{-1.1em} \textsuperscript{*} indicates equal contribution, \textsuperscript{\dag} indicates corresponding author.}

\authorrunning{R. Jin et al.}

% \institute{ Shanghai Jiao Tong University \and Shanghai AI Laboratory \and CUHK MMLab \and CPII under InnoHK \and Tsinghua University \\
% \email{renbiaojin@sjtu.edu.cn, guoshi@pjlab.org.cn}}
\institute{}
\maketitle

\begin{center}
{\small 
\textsuperscript{1}Shanghai Jiao Tong University \quad 
\textsuperscript{2}Shanghai AI Laboratory \quad   \\[0.3em]
\textsuperscript{3}CUHK MMLab \quad
\textsuperscript{4}CPII under InnoHK \quad 
\textsuperscript{5}Tsinghua University
}
\end{center}

\begin{abstract}
Real-world video deblurring remains challenging due to diverse motion patterns, complex degradations, and the scarcity of realistic training data, yet robust restoration is critical for downstream pipelines such as mobile imaging and 3D reconstruction. 
This work presents \textbf{RealVDeblur}, an efficient generative framework designed to improve in-the-wild robustness under diverse real capture conditions. 
First, a large-scale, physically grounded blur synthesis pipeline is constructed from scene-level 3D Gaussian Splatting (3DGS) assets and high-frame-rate videos, providing realistic training data covering both camera-induced and object-motion blur. 
Second, a video diffusion prior is leveraged for restoration; to better accommodate frame-dependent blur variations, temporal compression in the VAE is disabled and a frame-wise encoding scheme is adopted. 
For practical deployment on long videos, multi-step diffusion sampling is distilled into an efficient one-step generator, and a training-free Temporal Window Mask stabilizes inference beyond the training horizon with constant memory usage. 
Extensive experiments on diverse real-world benchmarks demonstrate strong perceptual quality, semantic fidelity, and temporal consistency on unseen videos, as well as improved robustness in downstream 3D reconstruction under severe motion blur. Project page: \href{https://rbjin.github.io/RealVDeblur/}{https://rbjin.github.io/RealVDeblur/}
\end{abstract}

\section{Introduction}
\label{sec:intro}

Real-world video deblurring is an important problem in computer vision, as motion blur frequently occurs in practical video capture due to camera shake, dynamic scenes, and long exposure. The task aims to recover sharp frames from blur degraded sequences. Video deblurring plays a critical role in downstream applications. For example, modern mobile imaging and video enhancement pipelines rely on stable sharp frames, and 3D reconstruction is particularly sensitive to motion blur, which significantly degrades geometric accuracy.

\begin{figure}[htbp]
  \centering
  \includegraphics[width=\textwidth]{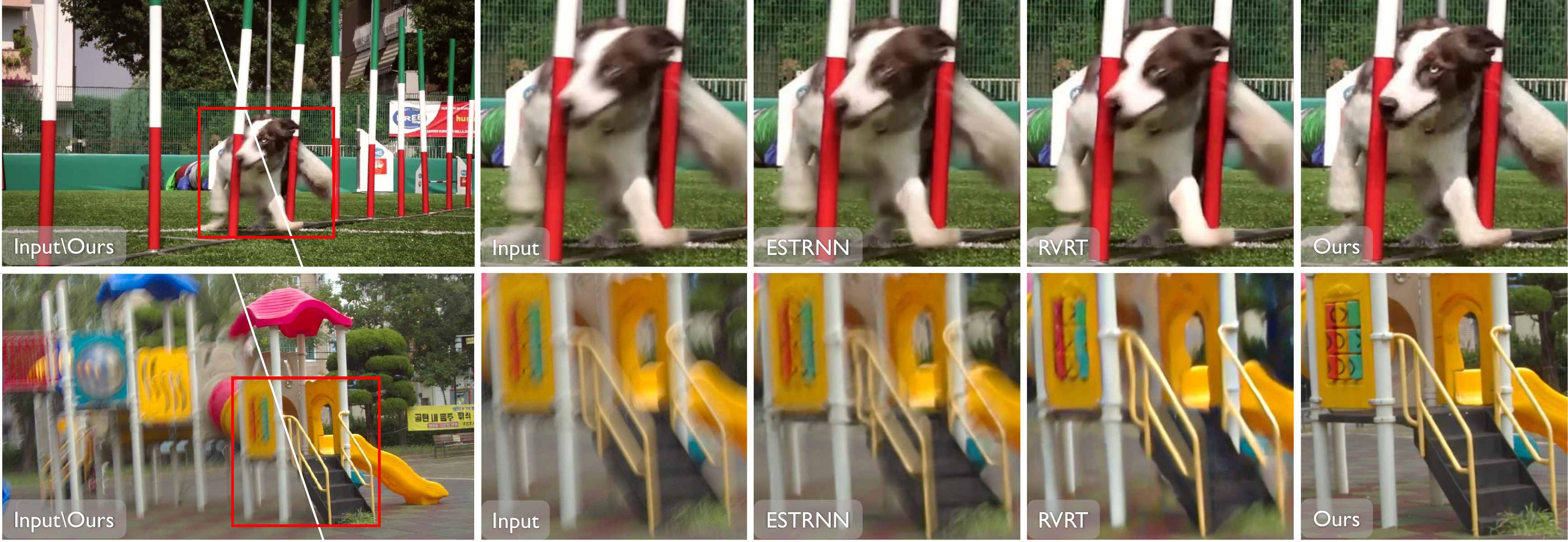}
  \caption{Visual comparison of our proposed \textbf{RealVDeblur} against state-of-the-art video deblurring methods in real-world scenarios. While traditional regression-based methods struggle with in-the-wild complex motion blur and produce over-smoothed results, our generative framework faithfully recovers sharp structures and high-frequency details.}

  \label{fig:teaser}
  \vspace{-20pt}
\end{figure}

Video deblurring methods have progressed from explicit motion alignment~\cite{pan2020cdvd_tsp,wang2019edvr} to implicit correspondence learning~\cite{zhou2019stfan,son2021pvdnet} and, more recently, transformer-based spatio-temporal aggregation~\cite{liang2022vrt,liang2022rvrt,zhang2024bsstnet}. Although these approaches achieve strong results on synthetic benchmarks, their performance often degrades on real-world videos, showing limited generalization and unsatisfactory deblurring quality. This limitation is mainly caused by two factors. (1) \textbf{Training data}: The available training data is limited in scale and often mismatched to real motion blur. Existing models are typically trained on datasets captured from a small number of scenes (e.g., GoPro contains only 20 scenes), providing insufficient diversity for learning robust motion statistics. Moreover, blur is commonly synthesized by frame averaging in the RGB domain, which introduces a discrepancy from the physical formation process of real motion blur. (2) \textbf{Lack of video prior}: Most prior methods adopt a deterministic regression formulation and therefore lack a realistic prior over sharp video sequences. Without such a prior, restored frames may exhibit residual blur and over-smoothed textures when applied to videos outside the training distribution. 

Consequently, developing a real-world video deblurring method that generalizes to diverse in-the-wild videos remains an open and challenging problem.

In this paper, we present \textbf{RealVDeblur}, an efficient generative framework for real-world video deblurring, designed to improve robustness under diverse capture conditions. Our framework addresses real-world generalization from both the data and modeling perspectives. By combining carefully constructed large-scale training data with a video diffusion prior and efficient inference design, RealVDeblur enables robust restoration under severe blur and diverse real capture conditions.

Existing video deblurring datasets are limited in scale and scene diversity (e.g., GoPro contains only 20 training scenes, and BSD provides around 60 training scenes), which restricts robust motion modeling. To address this limitation, we construct a large-scale realistic data generation pipeline inspired by real-world image blur simulation~\cite{rim2022realistic,lee2024gs}. We render synthetic sequences from scene-level 3D Gaussian Splatting (3DGS) assets~\cite{yeshwanth2023scannet++,interiorgs2025} and simulate camera-induced motion blur as well as defocus blur directly in the RAW domain through an ISP-aware process, enabling physically consistent degradation modeling. In addition, we incorporate large-scale high-frame-rate videos~\cite{nah2017gopro,nah2019reds,noki2025slomo} to better capture object-motion blur. In total, we curate $\sim$2,000 3DGS scenes and $\sim$3,000 high-frame-rate videos to produce diverse training data, significantly enhancing generalization to real-world capture conditions.

On the model side, to better model temporal dynamics and provide a realistic prior over sharp video sequences, we leverage recent advances in video diffusion models. While prior works such as VD-Diff~\cite{rao2024rethinking} and DIVD~\cite{long2024divd} demonstrate strong performance on synthetic benchmarks, directly applying pre-trained diffusion models to real-world blur remains challenging. Most video diffusion backbones rely on temporally compressed VAE latents, which assume smooth inter-frame transitions—an assumption often violated in real-world blurry sequences with varying blur magnitudes. Moreover, multi-step inference is computationally expensive and difficult to scale to long videos. In particular, positional encodings such as 3D RoPE introduce length extrapolation issues, leading to unstable inference when test sequences exceed the training horizon. To address these limitations, we introduce two key modifications. First, we disable temporal compression in the VAE and adopt a frame-wise 2D encoding scheme, enabling more faithful modeling of frame-level blur variations. Second, we distill the original multi-step diffusion process into an efficient one-step generator and introduce a training-free Temporal Window Mask that constrains the effective range of 3D RoPE to local temporal windows. 

This design alleviates length extrapolation issues and enables memory-efficient inference across long-range sequences, making diffusion-based restoration practical for long real-world videos.

Extensive experiments demonstrate that the proposed \textbf{RealVDeblur} generalizes across diverse real-world benchmarks, achieving strong perceptual quality, semantic fidelity, and temporal consistency on unseen datasets. Moreover, our method effectively mitigates motion and defocus blur in downstream 3D reconstruction scenarios, particularly under large-scale scene capture.

\noindent\textbf{Contributions.} Our contributions are summarized as follows.

\begin{itemize}[nosep, leftmargin=*]
\item We repurpose a pre-trained DiT-based video generation model as a generative spatio-temporal prior for video deblurring, with frame-wise latent encoding to model frame-dependent blur variations, enabling a unified model to handle camera-motion, object-motion, and defocus blur.
\item We develop a practical inference pipeline that combines single-step distribution matching distillation and a training-free temporal window strategy that eliminates RoPE extrapolation artifacts, enabling stable and efficient inference on long videos beyond the training horizon.
\item We provide an extensive multi-dimensional evaluation across diverse  real-world benchmarks, explicitly characterizing the perceptual-fidelity trade-off inherent in generative video restoration and establishing a strong baseline for VDM-based video deblurring.
\end{itemize}

\section{Related Work}

\subsection{Video Deblurring Networks}
Learning-based video deblurring methods aggregate complementary sharp evidence from neighboring frames via explicit or implicit alignment.
Explicit approaches estimate optical flow for feature warping~\cite{pan2020cdvd_tsp, wang2019edvr}, while implicit methods adopt filter-adaptive convolution~\cite{zhou2019stfan} or recurrent propagation with blur-robust compensation~\cite{son2021pvdnet, zhong2020estrnn} to avoid brittle motion estimation under heavy blur.
More recently, Transformer-based architectures such as VRT~\cite{liang2022vrt}, RVRT~\cite{liang2022rvrt}, and BSSTNet~\cite{zhang2024bsstnet} improve long-range spatio-temporal modeling.
Although these task-specific networks achieve strong distortion metrics, they are trained from scratch on limited paired data and tend to produce over-smoothed textures on complex real-world blur.
In contrast, we repurpose a pre-trained video generation model as a generative spatio-temporal prior to recover details beyond the regression capacity of conventional backbones.

\subsection{Blur Synthesis and Real-World Benchmarks}
Deblurring progress is closely tied to dataset realism. Real-world paired datasets have been collected with specialized rigs for images~\cite{rim2020real} and videos~\cite{zhong2023real}.
To narrow the synthetic-to-real gap, RSBlur~\cite{rim2022realistic} proposes a more realistic synthesis pipeline, and RAW-Blur~\cite{zhong2023real} advocates synthesizing blur in RAW space.
Event-assisted~\cite{kim2024frequency} and geometry-driven benchmarks such as GS-Blur~\cite{lee2024gs} further enrich evaluation coverage.
Following this trend, we curate training data covering camera motion, object motion, and defocus blur, and conduct multi-dimensional evaluations across diverse real-world benchmarks.

\subsection{Diffusion Priors and Video Foundation Models for Deblurring}
Diffusion models have been increasingly adopted as generative priors for low-level vision. 
For images, DiffBIR~\cite{lin2024diffbir} leverages latent diffusion for detail regeneration, and DeblurDiff~\cite{kong2025deblurdiff} customizes diffusion for real-world deblurring. For videos, VD-Diff~\cite{rao2024rethinking} integrates diffusion into a transformer backbone for restoration, and DIVD~\cite{long2024divd} adapts a video diffusion model with temporal attention for inter-frame dependency. Plug-and-play approaches apply diffusion priors without task-specific training but often suffer from temporal inconsistency and high costs~\cite{yang2024noise, yeh2024diffir2vr, li2025tdm}.

Large-scale video diffusion models~\cite{peebles2023dit, wan2025wan} encode rich statistics yet are expensive due to iterative sampling. Parameter-efficient adaptation~\cite{hu2022lora} enables task steering while retaining priors, and distillation techniques such as DMD~\cite{yin2024dmd} and progressive distillation~\cite{salimans2022progressive} reduce sampling to few or single steps. For long sequences, existing frameworks rely on windowed processing or sparse attention~\cite{liang2022vrt,liang2022rvrt,chen2026anyrecon}. Our method combines condition injection with LoRA to repurpose video foundation models, distills multi-step sampling into one-step inference via DMD, and introduces training-free temporal window masking for stable long-video inference.

\begin{figure}[t]
  \centering
  \includegraphics[width=\textwidth]{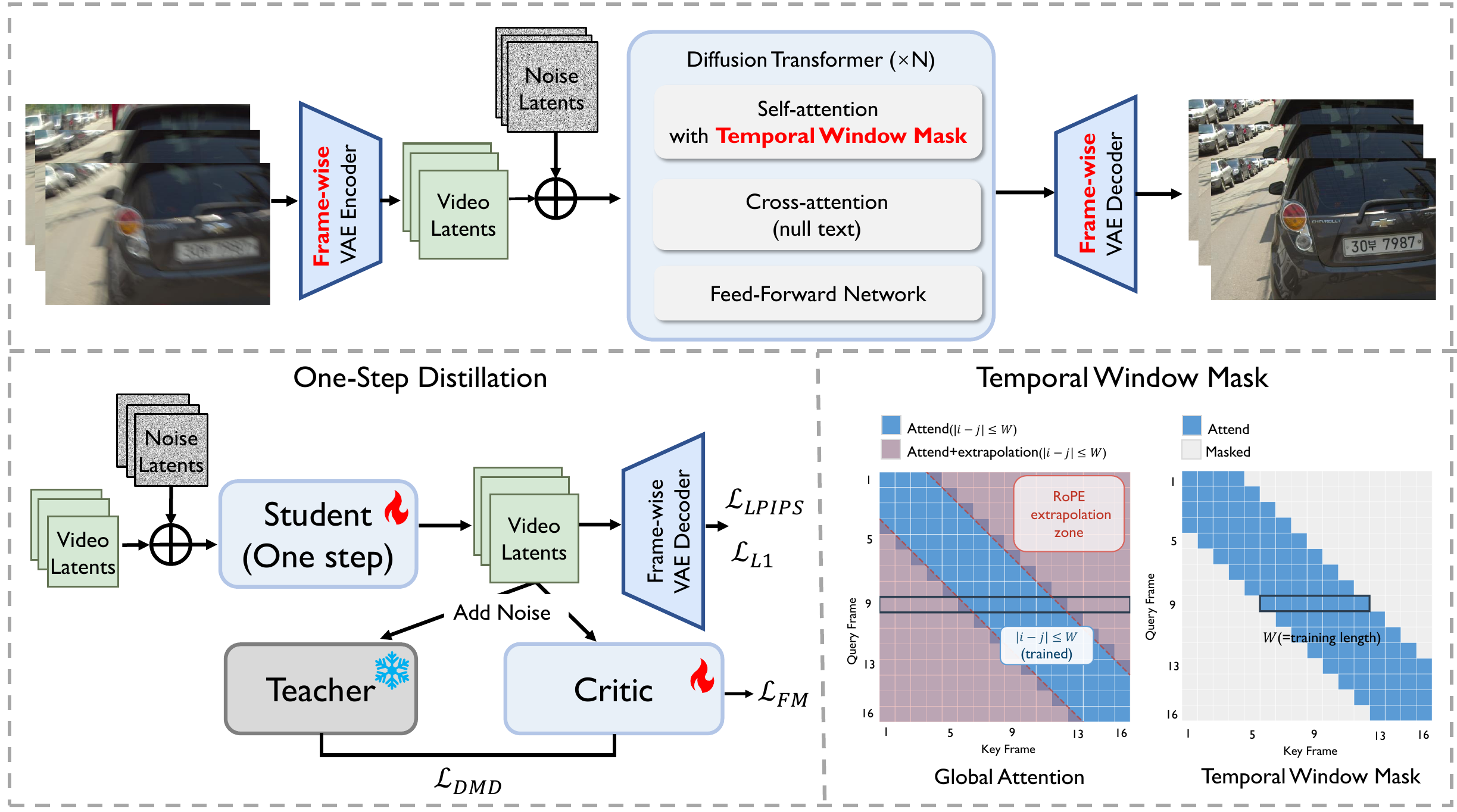}
  \caption{Overview of the \textbf{RealVDeblur} framework. Our approach repurposes a pre-trained Video Diffusion Model (VDM) for video deblurring with three key components: (1) a frame-wise VAE encoding scheme that preserves frame-dependent blur variations, (2) one-step inference via Distribution Matching Distillation (DMD) with pixel-space supervision, and (3) a training-free Temporal Window Mask that restricts self-attention to local windows, eliminating RoPE extrapolation instability and enabling constant-memory inference on arbitrarily long videos.}
  \label{fig:overall_architecture}
 \vspace{-10pt}
\end{figure}

\section{Method}
\label{sec:method}
\vspace{-7pt}

We present \textbf{RealVDeblur}, a generative video deblurring framework built upon a pre-trained Video Diffusion Model (VDM), coupled with carefully designed training data to enable robust in-the-wild deblurring under diverse real capture conditions. The overall pipeline is illustrated in Fig.~\ref{fig:overall_architecture}. 

Our framework repurposes the pre-trained VDM for video deblurring by injecting blurry observations as conditioning signals. To handle frame-dependent blur magnitudes that break the smooth-transition assumption of temporally compressed VAE latents, it disables temporal compression and adopts frame-wise encoding for faithful frame-level conditioning (Sec.~\ref{subsec:vdm_adaptation}). For efficient deployment, the multi-step sampling is distilled into a one-step model via distribution matching. A training-free temporal window strategy further stabilizes inference on arbitrarily long videos beyond the training horizon (Sec.~\ref{subsec:efficient_inference}). A comprehensive data curation and blur synthesis pipeline finally supplies diverse training samples spanning multiple blur types, improving in-the-wild generalization (Sec.~\ref{subsec:data_synthesis}).

\subsection{Adapting a Video Diffusion Model for Deblurring}
\label{subsec:vdm_adaptation}

We build upon Wan2.1~\cite{wan2025wan}, a large-scale Diffusion Transformer (DiT)~\cite{peebles2023dit} trained via Flow Matching~\cite{lipman2022flow} on large-scale image and video datasets. The model operates on latent representations produced by a Causal 3D Variational Autoencoder (Wan-VAE)~\cite{wan2025wan}.
Given a video $V \in \mathbb{R} ^ {T \times 3 \times H \times W}$, it is first encoded into a compact latent representation $z$, and then patchified into a sequence of latent tokens. The DiT processes these tokens through a stack of transformer blocks, each comprising spatio-temporal self-attention, cross-attention for text conditioning, and a feed-forward network. In this work, we adapt this pre-trained architecture for the video deblurring task.

\noindent\textbf{Conditioning and Parameter-Efficient Fine-Tuning.}
To enable conditional video deblurring, the blurry latent $z_{\mathrm{blur}}$ is projected to the latent feature space using a 3D convolutional network. 
The resulting embedding is added to the noisy latent $x_\tau$ before being fed into the DiT. 
Parameter-efficient fine-tuning is performed using LoRA modules attached to the transformer blocks, while the pre-trained VAE remains frozen. Text guidance is disabled via a null text embedding. Training follows the flow matching objective:
\begin{equation}
  \mathcal{L}_{\mathrm{FM}}
  = \mathbb{E}_{\tau,\epsilon}\,
    \bigl\lVert
      v_\theta(x_{\tau},\tau,\,z_{\mathrm{blur}}) - u_{\tau}
    \bigr\rVert_2^2,
\end{equation}
where $x_{\tau} = (1-\tau)x_0 + \tau\epsilon$ is the interpolated latent at time $\tau\in[0,1]$, and $u_{\tau} = \epsilon - x_0$ is the target velocity.

\noindent\textbf{Frame-wise Latent Encoding.}
Modern video diffusion models commonly employ 3D VAEs to jointly compress videos along spatial and temporal dimensions for improved efficiency. For example, Wan-VAE performs temporal downsampling (e.g., $\times4$) to obtain compact spatio-temporal latents for diffusion modeling~\cite{wan2025wan}. However, such temporal compression is less suitable for video restoration, particularly motion deblurring. Motion blur integrates scene content over time, causing large differences between adjacent frames and making temporal compression in the VAE prone to information loss.

To address this issue, we remove temporal compression in the VAE and instead adopt a frame-wise 2D encoder. For an input video $V=\{y_t\}_{t=1}^{T}$, we encode each frame independently and concatenate the resulting latents along the temporal dimension:
\begin{equation}
z_{\mathrm{blur}} =
\mathrm{concat}\bigl(E(y_1),\ldots,E(y_T)\bigr)\in\mathbb{R}^{T\times C\times H'\times W'},
\end{equation}
where $E$ denotes the frame-wise 2D VAE encoder. Although the VAE is modified to remove temporal compression, this change does not introduce a latent distribution mismatch for the pre-trained DiT. Wan2.1 is trained on both image and video datasets, where images can be regarded as videos with a temporal length of one. Therefore, the pre-trained DiT naturally supports frame-wise latent representations.

\subsection{Efficient Video Deblurring Inference}
\label{subsec:efficient_inference}

Deploying the multi-step VDMs for practical video deblurring faces two efficiency challenges: the high latency of iterative sampling (e.g., 50 denoising steps) and the poor scalability of global attention with RoPE extrapolation on long sequences. We address these issues with one-step distillation and a training-free temporal window mask.

\noindent\textbf{One-Step Distillation.}
To reduce the latency of iterative sampling, we adopt Distribution Matching Distillation (DMD)~\cite{yin2024dmd} to distill the multi-step diffusion process (e.g., 50 steps) into a one-step generator. DMD trains a Student model to match the output distribution of a frozen Teacher via the score difference between the Teacher and a learned Critic.
In our implementation, the Student, Teacher, and Critic share the same frozen DiT backbone and differ only in their independently trainable LoRA parameters. Starting from pure noise $z$, the Student directly predicts the clean latent $\hat{x}_0$ in a single step. The Student is trained using the DMD objective together with pixel-space $\ell_1$ and LPIPS~\cite{zhang2018lpips} losses computed on VAE-decoded frames: $\mathcal{L}_{\text{student}} = \mathcal{L}_{\text{DMD}} + \lambda_{L1}\mathcal{L}_{L1} + \lambda_{LPIPS}\mathcal{L}_{LPIPS}$
% \begin{equation}
% \mathcal{L}_{\text{student}} =
% \mathcal{L}_{\text{DMD}} +
% \lambda_{L1}\mathcal{L}_{L1} +
% \lambda_{LPIPS}\mathcal{L}_{LPIPS}.
% \end{equation}
The Critic is trained on the Student's detached outputs using a standard flow matching objective.
Empirically, we observe that the additional pixel-space supervision introduced during distillation also improves deblurring performance, as shown in the Sec.~\ref{sec:ablation}.

\noindent\textbf{Training-Free Temporal Window Mask.}
For video deblurring, each frame's restoration primarily depends on the temporal redundancy from nearby frames rather than global context across the entire sequence. However, directly applying VDMs to long video sequences leads to two major challenges: (1) the quadratic memory cost of global self-attention, and (2) the performance degradation of RoPE extrapolation when processing sequences longer than those seen during training~\cite{su2024roformer,wei2025videorope}.
We address these challenges with a \textbf{Temporal Window Mask}. During inference, we replace global self-attention with local window attention: for each frame $t$, attention is restricted to a temporal window of size $W$:
\begin{equation}
    \text{Attn}(Q_t, K, V) = \text{Softmax}\left( \frac{Q_t K_{[t - W/2, t + W/2]}^T}{\sqrt{d}} \right) V_{[t - W/2, t + W/2]}
\end{equation}
By constraining the attention to a fixed window $W$ that matches the training sequence length, we ensure that the relative distances between frames during inference remain strictly within the well-trained manifold of the rotary embeddings. This alignment between the inference and training positional distributions effectively eliminates the instability caused by RoPE extrapolation. Consequently, the model can maintain a consistent restoration quality for extended sequences by focusing exclusively on the most relevant local temporal context, without suffering from the cumulative errors typically observed in long-range global attention.

\subsection{Large-Scale Realistic Blur Synthesis}
\label{subsec:data_synthesis}
Real-world video blur arises from three physical mechanisms: camera shake, object motion, and optical defocus. However, existing training datasets cover only a subset of these degradations and are drawn from a limited number of scenes (e.g., 20 for GoPro, 60 for BSD), restricting the diversity of motion patterns and scene content available for training.
To address this, we build a large-scale training set, termed \textbf{OmniBlur}, by synthesizing realistic blur from two complementary data sources:
(1) 3D Gaussian Splatting (3DGS) reconstructions~\cite{kerbl20233dgs} for camera-motion and defocus blur, and (2) high-frame-rate videos~\cite{nah2017gopro,nah2019reds,noki2025slomo} for object-motion blur.

\noindent\textbf{Camera-Motion and Defocus Blur from 3DGS.}
Following GS-Blur~\cite{lee2024gs}, which shows that rendering along perturbed camera poses in a 3DGS scene can produce realistic image-level motion blur, we extend this idea to the video setting using large-scale 3DGS scenes~\cite{yeshwanth2023scannet++,interiorgs2025}.
For camera-motion blur, we generate a smooth camera trajectory through each scene and add random 6-DoF perturbations via B\'{e}zier curves to mimic hand-held shake; $K$ sub-frames are rendered and averaged in linear space to produce the blurry frame.
For defocus blur, we use a thin-lens camera model with per-pixel depth from 3DGS rendering: $N$ views are sampled on a circular aperture disk around a randomly chosen focal plane, producing realistic depth-dependent bokeh.
Since camera shake and defocus often co-occur in practice, we also generate compound blur by combining both mechanisms ($K\!\times\!N$ renderings per frame).

\noindent\textbf{Object-Motion Blur from Dynamic Videos.}
Camera-motion and defocus blur can be rendered from static 3DGS scenes, but object-motion blur requires real dynamic content. We incorporate three high-frame-rate datasets (GoPro~\cite{nah2017gopro}, REDS~\cite{nah2019reds}, and SloMo~\cite{noki2025slomo}) to provide coverage of large object displacements and non-rigid motion.

\noindent\textbf{ISP-Aware Augmentation and Data Composition.}
Rendered images are noise-free and skip the camera ISP, creating a gap with real captures. Following RSBlur~\cite{rim2022realistic}, we inject Poisson--Gaussian noise in the RAW domain and simulate the ISP (mosaicing, white balance, demosaicing) to produce realistic training pairs.
Our resulting OmniBlur contains $\sim$2,000 3DGS scenes and $\sim$3,000 high-frame-rate videos. 
To improve robustness across diverse capture conditions, we adopt a mixed-training strategy that incorporates varying sequence lengths, multiple spatial resolutions, and various blur types (camera-motion, object-motion, and defocus) during training. Details of the dataset are provided in the supplementary material.

\section{Experiments}
\label{sec:experiments}
\subsection{Experimental Setup}
\noindent\textbf{Benchmarks.}
To evaluate the real-world generalization capability of \textbf{RealVDeblur}, we focus on diverse real-world benchmarks rather than synthetic ones like GoPro~\cite{nah2017gopro} and DVD~\cite{su2017dvd}. Specifically, we conduct evaluations on BSD~\cite{zhong2023real}, RealBlur~\cite{rim2020real}, RSBlur~\cite{rim2022realistic}, FEVD~\cite{kim2024frequency}, and RWBI~\cite{zhang2020rwbi}. We extract 50 and 165 sequences from the test sets of RealBlur and RSBlur, respectively, with an average length of 20 frames per sequence. For BSD, we utilize the 3ms–24ms test subset, comprising 20 sequences of 150 frames each. For FEVD, we employ the RGB sequences from the 8 test videos of 150 frames each, and for RWBI, we use 43 reorganized sequences with lengths ranging from 10 to 100 frames. 

\noindent\textbf{Evaluation Metrics.}
We adopt a comprehensive set of metrics to evaluate restoration fidelity, perceptual quality, semantic realism, and temporal coherence. Specifically, PSNR and SSIM measure distortion-level reconstruction accuracy, while LPIPS~\cite{zhang2018lpips} and FID~\cite{heusel2017fid} assess perceptual similarity and distribution alignment. We further report no-reference quality metrics, including MUSIQ~\cite{ke2021musiq} and NIQE~\cite{mittal2012niqe}, to reflect high-level visual quality. Temporal consistency is evaluated using temporal optical flow error (tOF~\cite{chu2020learning}).

\noindent\textbf{Baseline.}
We compare RealVDeblur with representative recurrent-based methods (ESTRNN~\cite{zhong2020estrnn}, RNN-MBP~\cite{zhu2022deep}), alignment-based CNN approaches (ShiftNet~\cite{li2023shiftnet}), and transformer-based models (VRT~\cite{liang2024vrt}, RVRT~\cite{liang2022rvrt}, BSSTNet~\cite{zhang2024bsstnet}). Among them, ESTRNN is trained on the real-world BSD dataset, while the remaining methods are trained on synthetic datasets such as GoPro and REDS. For VRT, we adopt the REDS-trained model as it achieves better performance than its GoPro-trained counterpart. 
% For 3DGS evaluation, we further compare with BAGS, Deblurring-3DGS, Restormer+3DGS, and Turtle+3DGS.

\noindent\textbf{Implementation Details.}
We fine-tune Wan2.1-1.3B with LoRA (rank 64) using AdamW with 
learning rate 5e-5 on 32$\times$A100 GPUs. DMD distillation uses $\lambda_{L1}\!=\lambda_{LPIPS}\!=\!2$. The TWM window size is set to $W\!=\!20$, matching the training sequence length. Full details are in the supplement.

\begin{table}[htbp]
    \centering
    \caption{Quantitative comparison on real-world video deblurring benchmarks. Best and second-best results are \textbf{bolded} and \underline{underlined}, respectively.}
        \label{tab:main_results}
    \resizebox{\textwidth}{!}
    {
    \begin{tabular}{lll ccccccc}
        \toprule
        \multirow{2}{*}{\textbf{Benchmark}} & \multirow{2}{*}{\textbf{Method}} & \multirow{2}{*}{\textbf{\makecell{Training \\ Dataset}}} & \multicolumn{2}{c}{\textbf{Distortion}} & \multicolumn{2}{c}{\textbf{Perceptual}} & \multicolumn{2}{c}{\textbf{Semantic}} & \textbf{Temporal} \\
        \cmidrule(lr){4-5} \cmidrule(lr){6-7} \cmidrule(lr){8-9} \cmidrule(lr){10-10}
        & & & PSNR$\uparrow$ & SSIM$\uparrow$ & LPIPS$\downarrow$ & FID$\downarrow$ & MUSIQ$\uparrow$ & NIQE$\downarrow$ & tOF$\downarrow$ \\
        \midrule
        \multirow{7}{*}{\makecell{BSD  ~\cite{zhong2023real}}}
        & ESTRNN\cite{zhong2020estrnn} & BSD & \textbf{31.39} & \textbf{0.936} & \textbf{0.114} & \underline{10.8} & \underline{43.88} & 5.793 & \underline{2.146} \\
        & RNN-MBP\cite{zhu2022deep} & GoPro & 22.84 & 0.726 & 0.320 & 69.6 & 33.05 & \underline{5.426} & 6.639 \\
        & ShiftNet\cite{li2023shiftnet} & GoPro & 24.98 & 0.816 & 0.221 & 36.6 & 39.21 & 5.770 & 4.452 \\
        & VRT\cite{liang2024vrt} & REDS & 26.72 & 0.850 & 0.199 & 33.5 & 39.34 & 5.605 & 2.583 \\
        & RVRT\cite{liang2022rvrt} & GoPro & 26.06 & 0.849 & 0.196 & 28.6 & 39.72 & 5.563 & 3.725 \\
        & BSSTNet\cite{zhang2024bsstnet} & GoPro & 20.90 & 0.707 & 0.320 & 60.5 & 38.16 & 7.370 & 6.532 \\
        \rowcolor{gray!10} \cellcolor{white}&\textbf{RealVDeblur} & \textbf{OmniBlur} & \underline{28.76} & \underline{0.884} & \underline{0.118} & \textbf{9.1} & \textbf{52.49} & \textbf{4.583} & \textbf{1.974} \\

        \midrule
        \multirow{7}{*}{\makecell{RealBlur ~\cite{rim2020real}}}
        & ESTRNN\cite{zhong2020estrnn} & BSD    & \underline{26.90} & 0.825  & 0.207 & 32.3 & 41.64 & 6.050 & 2.469 \\
        & RNN-MBP\cite{zhu2022deep} & GoPro & 24.43             & 0.775 & 0.258 & 59.9 & 40.36 & 6.164 & 3.274 \\
        & ShiftNet\cite{li2023shiftnet} & GoPro & 23.95 & 0.762 & 0.198 & 33.7 & 49.44 & 5.408 & 4.423 \\
        & VRT\cite{liang2024vrt} & REDS & 26.54 & \underline{0.838} & 0.153 & 20.9 & 51.09 & 4.938 & 2.338 \\
        & RVRT\cite{liang2022rvrt} & GoPro & 26.00 & 0.839 & \underline{0.134} & \underline{14.8} & \underline{53.02}  & \underline{4.906} & \underline{2.206} \\
        & BSSTNet\cite{zhang2024bsstnet} & GoPro & 25.09 & 0.807 & 0.184 & 28.1 & 48.12 & 5.407 & 3.092 \\
        \rowcolor{gray!10} \cellcolor{white}&\textbf{RealVDeblur} & \textbf{OmniBlur} & \textbf{27.55} & \textbf{0.852} & \textbf{0.100} & \textbf{11.3} & \textbf{57.01} & \textbf{4.433} & \textbf{1.827} \\

        \midrule
        \multirow{7}{*}{\makecell{RSBlur ~\cite{rim2022realistic}}}
        & ESTRNN\cite{zhong2020estrnn} & BSD &\underline{30.26} &\underline{0.868} &0.290 &\underline{20.9} &34.70 &6.448 &\underline{3.240} \\
        & RNN-MBP\cite{zhu2022deep} & GoPro &25.79 &0.768 &0.352 &60.5 &29.59 &6.492 &9.605 \\
        & ShiftNet\cite{li2023shiftnet} & GoPro &25.75 &0.766 &0.326 &48.7 &\underline{39.10} &\underline{5.151} &14.59 \\
        & VRT\cite{liang2024vrt} & REDS &26.46 &0.824 &0.306 &48.0 &34.00 &5.674 &5.285 \\
        & RVRT\cite{liang2022rvrt} & GoPro &28.53 &0.854 &\underline{0.244} &27.7 &38.72 &5.444 &4.766 \\
        & BSSTNet\cite{zhang2024bsstnet} & GoPro &28.03 &0.830 &0.277 &35.2 &34.61 &6.024 &8.043 \\
        \rowcolor{gray!10} \cellcolor{white}&\textbf{RealVDeblur} & \textbf{OmniBlur} &\textbf{32.70} &\textbf{0.915} &\textbf{0.142} &\textbf{5.8} &\textbf{52.01} &\textbf{4.884} &\textbf{1.301} \\

        \midrule
        \multirow{7}{*}{\makecell{FEVD ~\cite{kim2024frequency}}}
        & ESTRNN\cite{zhong2020estrnn} & BSD & \underline{28.76} & \textbf{0.896} & \underline{0.219} & \underline{39.0} & \underline{29.57} & 8.076 & \underline{7.630} \\
        & RNN-MBP\cite{zhu2022deep} & GoPro & 25.75 & 0.843 & 0.302 & 89.0 & 26.58 & 7.611 & 16.14 \\
        & ShiftNet\cite{li2023shiftnet} & GoPro & 25.50 & 0.846 & 0.317 & 94.6 & 26.56 & 7.274 & 21.68 \\
        & VRT\cite{liang2024vrt} & REDS & 27.01 & 0.875 & 0.265 & 83.3 & 28.54 & \underline{6.840} & 11.85 \\
        & RVRT\cite{liang2022rvrt} & GoPro & 27.65 & 0.881 & 0.247 & 76.7 & 28.97 & 7.475 & 13.48 \\
        & BSSTNet\cite{zhang2024bsstnet} & GoPro & 25.43 & 0.846 & 0.269 & 77.4 & 28.41 & 7.465 & 15.83 \\
        \rowcolor{gray!10} \cellcolor{white}&\textbf{RealVDeblur} & \textbf{OmniBlur} & \textbf{28.90} & \underline{0.886} & \textbf{0.137} & \textbf{12.4} & \textbf{40.25} & \textbf{6.058} & \textbf{3.978} \\
        \bottomrule
    \end{tabular}
    }
\end{table}

\subsection{Evaluation on Real-world Video Deblurring}
\noindent\textbf{Quantitative Analysis.}
Table~\ref{tab:main_results} summarizes quantitative comparisons across four real-world benchmarks: BSD, RealBlur, RSBlur, and FEVD. RealVDeblur achieves the best or second-best results on nearly all metrics across all benchmarks, demonstrating strong cross-dataset generalization.
We highlight three key findings that corroborate our design motivations.
(1) Methods trained on synthetic datasets (e.g., BSSTNet, RNN-MBP on GoPro) suffer notable degradation on real-world benchmarks. This confirms the domain gap between synthetic frame-averaging blur and real motion blur, validating our physically grounded data construction pipeline.
(2) ESTRNN, trained on the real-world BSD dataset, achieves the highest PSNR on BSD (31.39) but consistently underperforms in perceptual metrics (e.g., MUSIQ 43.88 vs.\ our 52.49). This suggests that deterministic regression without a video prior tends to produce over-smoothed textures, supporting our adoption of a diffusion-based generative framework.
(3) RealVDeblur achieves the best temporal consistency (tOF) on all four benchmarks, demonstrating that our distilled one-step inference with temporal window mask effectively maintains inter-frame coherence.
On the RWBI benchmark Table~\ref{tab:rwbi_results}, RealVDeblur surpasses all baselines by a significant margin (MUSIQ: 63.50 vs.\ 47.85, CLIP-IQA: 0.510 vs.\ 0.356), further confirming the real-world generalization capability enabled by our large-scale training data and generative prior.

\begin{figure}[htbp]
  \centering
  \includegraphics[width=\textwidth]{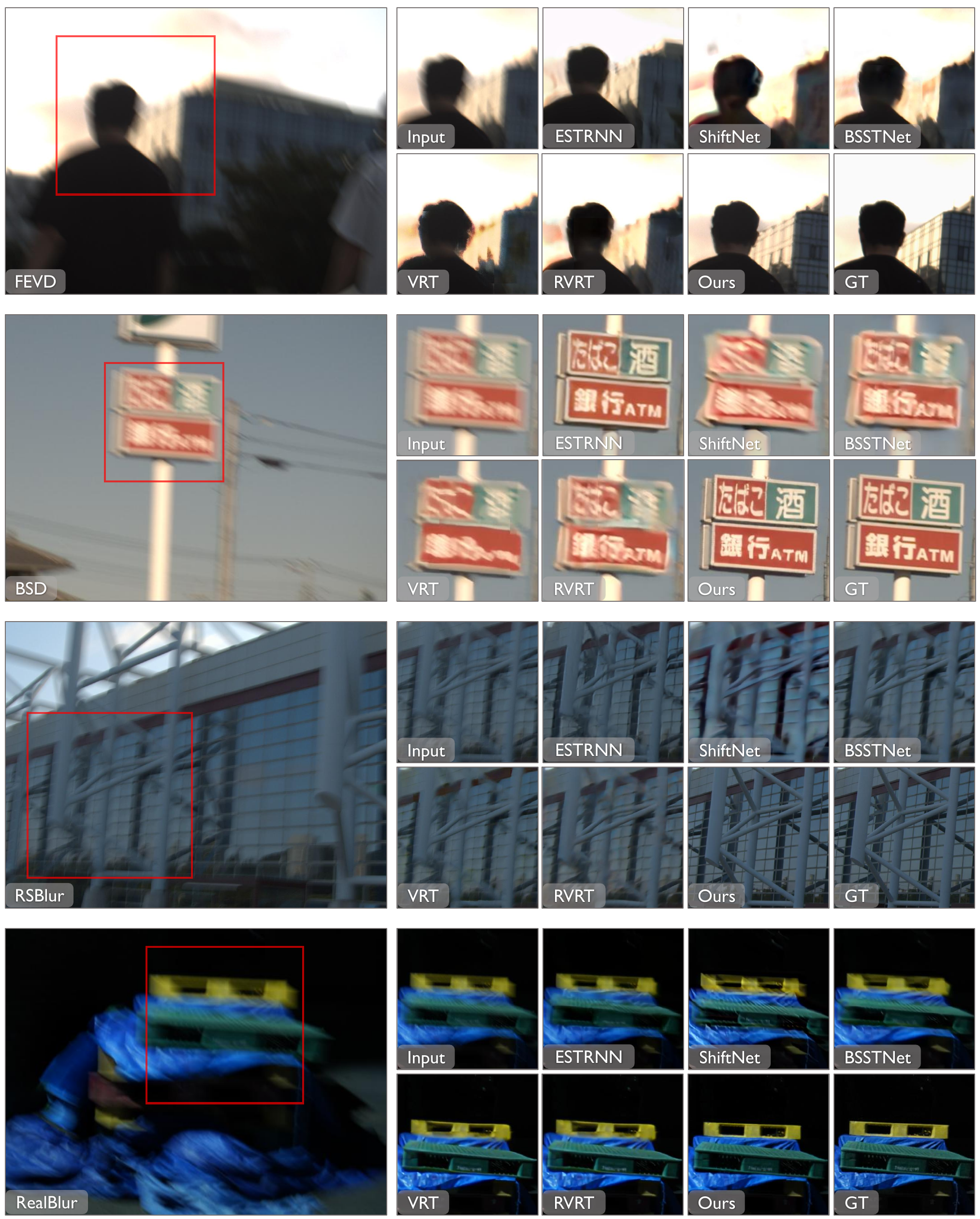}
  \caption{Qualitative comparison on multiple real-world evaluation datasets. Under challenging conditions including severe motion blur, low light, and overexposure, baseline models typically yield heavily smoothed outputs or structural distortions. In contrast, RealVDeblur(Ours) effectively recovers sharp, high-frequency details while maintaining geometric fidelity. GT denotes ground truth.}

  \label{fig:videodeblur}
\end{figure}

\noindent\textbf{Qualitative Analysis.}
 As shown in Figure~\ref{fig:videodeblur}, we present qualitative comparisons under challenging real-world conditions including severe motion blur in low-light and overexposed environments. Existing baselines struggle to reconstruct fine structural details; for instance, in the severely degraded FEVD examples, background architectural details are entirely lost due to overexposure and heavy blur, with all baselines producing heavily smoothed or distorted outputs. In contrast, RealVDeblur effectively leverages its learned generative prior and temporal modeling to recover high-frequency details that are missing from the blurry input. Furthermore, our method faithfully restores straight geometric structures (e.g., glass walls in RSBlur), whereas prior methods produce notable structural distortions, demonstrating the advantage of our diffusion-based approach in preserving geometric consistency.

\begin{table}[htbp]
  \centering
  \caption{Quantitative evaluation on the reference-free RWBI benchmark~\cite{zhang2020rwbi}.}
  \label{tab:rwbi_results}
  \resizebox{0.65\textwidth}{!}{
  \begin{tabular}{llccc}
    \toprule
    \textbf{Method} & \textbf{Data} & \textbf{MUSIQ$\uparrow$} & \textbf{CLIP-IQA$\uparrow$} & \textbf{NIQE$\downarrow$} \\
    \midrule
    ESTRNN\cite{zhong2020estrnn} & BSD & \underline{47.85} & \underline{0.356} & 5.759 \\
    RNN-MBP\cite{zhu2022deep} & GoPro & 38.48 & 0.207 & 5.553 \\
    ShiftNet\cite{li2023shiftnet} & GoPro & 44.14 & 0.244 & 5.837 \\
    VRT\cite{liang2024vrt} & REDS & 45.87 & 0.250 & \underline{4.913} \\
    RVRT\cite{liang2022rvrt} & GoPro & 45.59 & 0.257 & 5.356 \\
    BSSTNet\cite{zhang2024bsstnet} & GoPro & 43.58 & 0.215 & 5.774 \\
    \textbf{RealVDeblur} & \textbf{OmniBlur} & \textbf{63.50} & \textbf{0.510} & \textbf{4.250} \\
    \bottomrule
  \end{tabular}
  }
\end{table}

%--------------------------------------------------------------
\subsection{Evaluation on 3DGS Reconstruction}
\label{sec:real_world}

To validate the practical value of video deblurring for downstream 3D vision tasks, we evaluate its impact as a preprocessing step for 3D Gaussian Splatting (3DGS~\cite{kerbl20233dgs}) reconstruction. We compare against two end-to-end deblurring-3DGS methods—BAGS~\cite{peng2024bags} and Deblurring-3DGS~\cite{lee2024deblurring3dgs}—that jointly optimize blur modeling and scene representation, as well as cascaded baselines that apply VRT~\cite{liang2022vrt} for multi view image restoration before standard 3DGS training. Detailed experimental settings are provided in the supplementary material.

Overall, these results demonstrate that RealVDeblur serves as a strong general-purpose preprocessing module for 3DGS pipelines, consistently outperforming end-to-end alternatives on camera-motion blur and achieving competitive performance on defocus blur without requiring any task-specific joint optimization.

\begin{figure}[htbp]
  \centering
  \includegraphics[width=\textwidth]{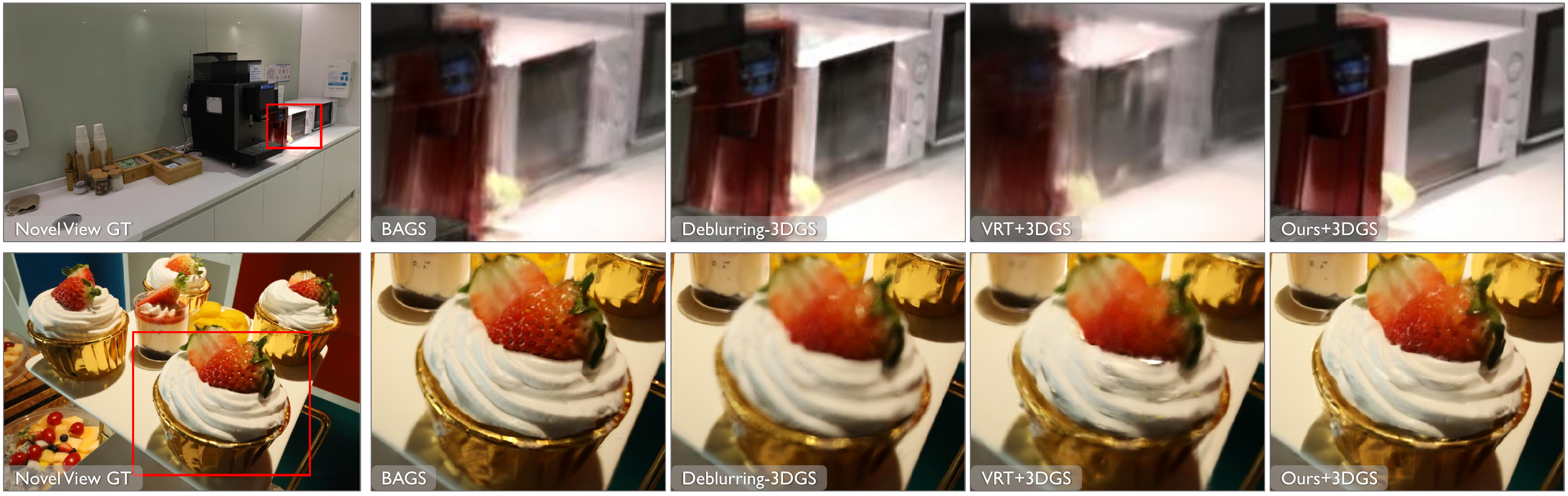}
  \caption{Visual evaluation of 3DGS reconstruction under real-world camera-motion blur (first row) and defocus blur (second row). Compared with end-to-end methods (BAGS, Deblurring-3DGS) and the cascaded baseline (VRT + 3DGS), using RealVDeblur as a preprocessing step substantially improves geometric accuracy and texture sharpness of the reconstructed scenes.}
  \label{fig:3d}
\end{figure}

\vspace{-15pt}

\begin{table}[htbp]
    \centering
    \caption{
    Quantitative comparison of 3DGS reconstruction quality on the DeblurNeRF\cite{ma2022deblur} dataset under real camera-motion and defocus blur. Best and second-best results are bolded and \underline{underlined}, respectively.}
    \label{tab:deblur-comparison}
    \footnotesize
    \renewcommand{\arraystretch}{1.2}
    \resizebox{0.75\textwidth}{!}{
    \begin{tabular}{l ccc ccc}
        \toprule
        \multirow{2}{*}{\textbf{Method}} & \multicolumn{3}{c}{\textbf{Real Camera Motion}} & \multicolumn{3}{c}{\textbf{Real Defocus}}  \\
        \cmidrule(lr){2-4} \cmidrule(lr){5-7}
        & PSNR $\uparrow$ & SSIM $\uparrow$ & LPIPS $\downarrow$ & PSNR $\uparrow$ & SSIM $\uparrow$ & LPIPS $\downarrow$ \\
        \midrule
        BAGS\cite{peng2024bags}             & 24.98 & 0.7458 & 0.2310 & \textbf{23.32} & 0.7242 & \underline{0.1846}\\
        Deblurring-3DGS\cite{lee2024deblurring3dgs}  & \textbf{25.67} & \underline{0.7695} & \underline{0.2231} & 22.81 & 0.6997 & 0.2236 \\
        \midrule
        VRT\cite{liang2024vrt} + 3DGS       & 22.56 & 0.7220 & 0.3225 & 21.85 & \underline{0.7873}& 0.2741 \\
        \textbf{RealVDeblur + 3DGS} & \underline{25.41} & \textbf{0.8095} & \textbf{0.1628} & \underline{22.87} & \textbf{0.8335} & \textbf{0.1798}\\
        \bottomrule
    \end{tabular}
    }
\end{table}

%--------------------------------------------------------------
\subsection{Ablation Studies}
\label{sec:ablation}
We conduct ablation studies on the BSD benchmark to validate each key design choice. Results are summarized in Table~\ref{tab:ablation}.

\begin{figure}[t!]
  \centering
  \includegraphics[width=\textwidth]{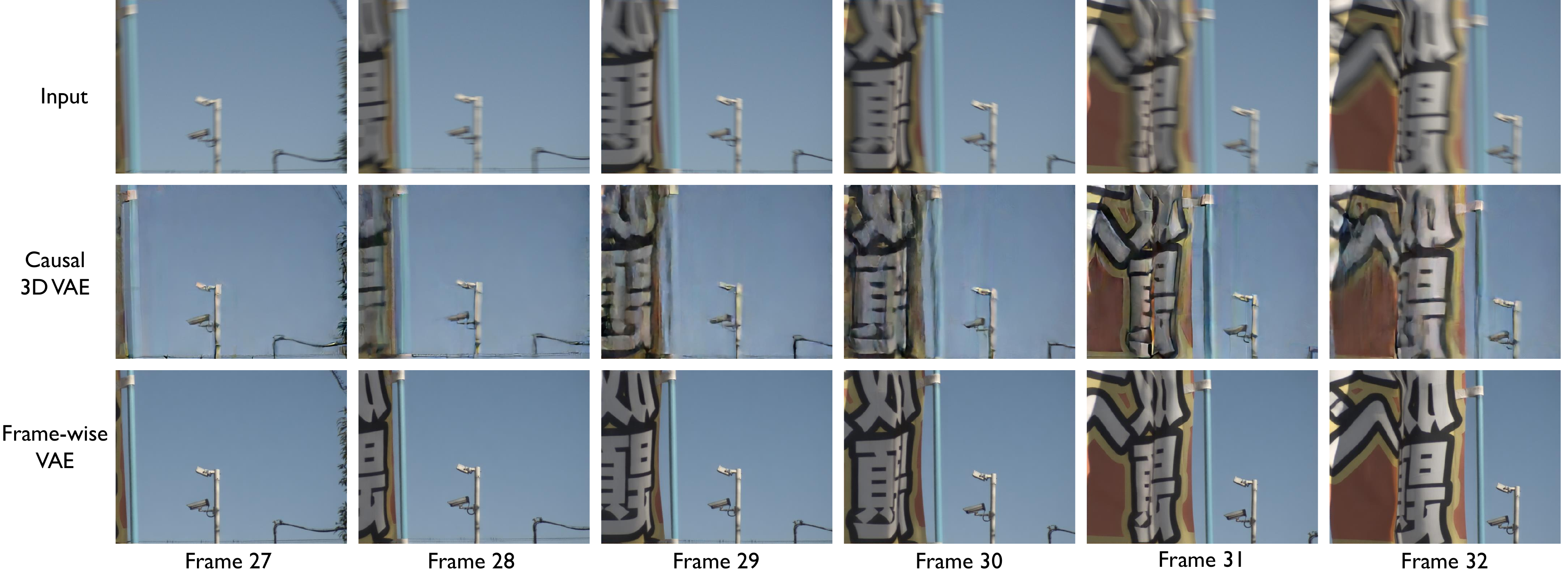}
  \caption{Comparison of Causal 3D VAE vs. Frame-wise VAE. In sequences with rapid motion (e.g., Frames 27-32), the Causal 3D VAE exhibits significant artifacts and temporal incoherence, while the Frame-wise VAE maintains sharp details and high reconstruction fidelity.}
  \label{fig:ab_vae}
\end{figure}

\noindent\textbf{Frame-wise VAE Encoding.}
Replacing our frame-wise encoder with the default causal 3D VAE causes the largest degradation among all ablations ($-$4.87 dB PSNR). The 3D VAE compresses adjacent frames into shared temporal latents, assuming smooth inter-frame transitions—an assumption violated by varying blur magnitudes across frames. This discrepancy is further amplified in the BSD dataset, where long exposures and rapid motions introduce extreme temporal misalignments. The resulting information loss prevents the DiT from modeling per-frame degradation, producing severe reconstruction artifacts and temporal incoherence. Fig.~\ref{fig:ab_vae} visualizes the deblurring performance of different VAEs in a sequence characterized by large motion.

\noindent\textbf{3DGS-Rendered Training Data.}
Removing 3DGS-rendered data and training solely on high-frame-rate videos leads to a 2.30 dB PSNR drop. High-frame-rate datasets primarily cover object-motion blur but lack camera ego-motion and defocus blur, which are prevalent in real-world captures. Our 3DGS pipeline provides complementary coverage of these degradation types with ISP-aware noise augmentation, substantially improving generalization.

\noindent\textbf{One-Step Distillation (DMD).}
Counter-intuitively, the distilled one-step model slightly outperforms the 50-step teacher. We attribute this to the pixel-space supervision ($\ell_1$ and LPIPS losses) introduced during distillation, which provides direct image-domain gradients absent in the original latent-space flow matching training. The distilled model thus achieves both better quality and faster inference time.

\begin{figure}[htbp]
  \centering
  \includegraphics[width=\textwidth]{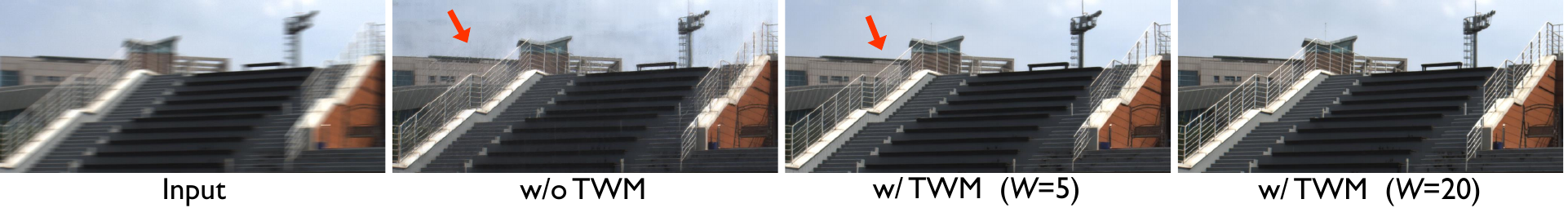}
  \caption{Qualitative ablation of the Temporal Window Mask on a 150-frame video across three attention configurations: global attention (w/o TWM), a narrow window ($W=5$), and our default window ($W=20$). Global attention causes severe artifacts due to RoPE extrapolation; a narrow window avoids extrapolation but limits temporal aggregation; our default window achieves the best balance between stability and restoration quality.}
  \label{fig:ab_twm}
  % \vspace{-20pt}
\end{figure}

\noindent\textbf{Temporal Window Mask (TWM).}
As shown in Figure~\ref{fig:ab_twm}, we qualitatively evaluate the TWM on a 150-frame video. Global attention (without TWM) causes severe artifacts due to RoPE extrapolation beyond the training horizon. A narrow window ($W=5$) avoids extrapolation but limits temporal aggregation, leaving residual blur. Our default window ($W=20$) matches the training length, balancing stable positional encoding with sufficient temporal context for high-fidelity restoration.

\begin{table}[htbp]
    \centering
    \caption{\textbf{Ablation of Proposed Components.} We evaluate the contribution of each component by replacing or removing it from our full pipeline (last row). All experiments are evaluated on the BSD benchmark.}
    \label{tab:ablation}
    \resizebox{0.9\textwidth}{!}{
    \begin{tabular}{l c c c c c}
        \toprule
        Variant & PSNR $\uparrow$ & SSIM $\uparrow$ & LPIPS $\downarrow$ & tOF $\downarrow$ & Time(s/frame) $\downarrow$ \\
        \midrule
        w/o Frame-wise VAE(Causal 3D VAE) & 23.89 & 0.748 & 0.262 & 9.106 & -\\
        w/o 3DGS-Rendered Data & 26.46 & 0.810 & 0.188 & 2.782 & - \\
        w/o DMD (50-step Teacher) & 28.02 & 0.873 & 0.135 & 2.097 & 3.53\\
        w/o TWM (Global Attention) & 27.20 & 0.845 & 0.165 & 3.786 & 0.29\\
        \midrule
        \textbf{RealVDeblur}(Full Model) & \textbf{28.76} & \textbf{0.884} & \textbf{0.118} & \textbf{1.974} & \textbf{0.15} \\
        \bottomrule
    \end{tabular}
    }
\end{table}
% \vspace{-20pt}

\vspace{10pt}

%\section{Limitations and Future Work}
%\label{sec:limitations}
\vspace{-10pt}

\section{Conclusion}
This paper presented \textbf{RealVDeblur}, an efficient diffusion-based framework for real-world video deblurring with strong cross-dataset generalization. RealVDeblur addresses the two major bottlenecks of in-the-wild deblurring—mismatched training data and the lack of a realistic video prior—by combining a large-scale, physically grounded blur synthesis pipeline with a pre-trained video diffusion prior. To better accommodate frame-dependent blur variations, we disable temporal compression in the VAE and adopt frame-wise encoding. For practical deployment, we distill multi-step diffusion sampling into a one-step generator and introduce a training-free Temporal Window Mask, enabling stable long-video inference beyond the training horizon with constant memory usage. Extensive experiments demonstrate improved perceptual quality and temporal consistency on diverse real-world benchmarks, and further validate the benefit of our deblurring results for downstream 3D reconstruction under severe motion blur.

%\section*{Acknowledgements}
%Please insert your acknowledgments here.

\bibliographystyle{splncs04}
\bibliography{main}

@String(CVPR  = {IEEE Conf. Comput. Vis. Pattern Recog.})

@String(ICLR  = {Int. Conf. Learn. Represent.})

@String(AAAI  = {AAAI})

@String(TOG   = {ACM Trans. Graph.})

@String(CVPR  = {CVPR})

@String(ICLR  = {ICLR})

@String(TOG   = {ACM TOG})

@inproceedings{pan2020cdvd_tsp,
  title={Cascaded deep video deblurring using temporal sharpness prior},
  author={Pan, Jinshan and Bai, Haoran and Tang, Jinhui},
  booktitle={Proceedings of the IEEE/CVF conference on computer vision and pattern recognition},
  pages={3043--3051},
  year={2020}
}

@inproceedings{wang2019edvr,
  title={Edvr: Video restoration with enhanced deformable convolutional networks},
  author={Wang, Xintao and Chan, Kelvin CK and Yu, Ke and Dong, Chao and Change Loy, Chen},
  booktitle={Proceedings of the IEEE/CVF conference on computer vision and pattern recognition workshops},
  pages={0--0},
  year={2019}
}

@inproceedings{zhou2019stfan,
  title={Spatio-temporal filter adaptive network for video deblurring},
  author={Zhou, Shangchen and Zhang, Jiawei and Pan, Jinshan and Xie, Haozhe and Zuo, Wangmeng and Ren, Jimmy},
  booktitle={Proceedings of the IEEE/CVF international conference on computer vision},
  pages={2482--2491},
  year={2019}
}

@article{son2021pvdnet,
  title={Recurrent video deblurring with blur-invariant motion estimation and pixel volumes},
  author={Son, Hyeongseok and Lee, Junyong and Lee, Jonghyeop and Cho, Sunghyun and Lee, Seungyong},
  journal={ACM Transactions on Graphics (TOG)},
  volume={40},
  number={5},
  pages={1--18},
  year={2021},
  publisher={ACM New York, NY}
}

@inproceedings{zhong2020estrnn,
  title={Efficient spatio-temporal recurrent neural network for video deblurring},
  author={Zhong, Zhihang and Gao, Ye and Zheng, Yinqiang and Zheng, Bo},
  booktitle={European conference on computer vision},
  pages={191--207},
  year={2020},
  organization={Springer}
}

@article{liang2022vrt,
  title={Vrt: A video restoration transformer},
  author={Liang, Jingyun and Cao, Jiezhang and Fan, Yuchen and Zhang, Kai and Ranjan, Rakesh and Li, Yawei and Timofte, Radu and Van Gool, Luc},
  journal={IEEE Transactions on Image Processing},
  volume={33},
  pages={2171--2182},
  year={2024},
  publisher={IEEE}
}

@article{liang2022rvrt,
  title={Recurrent video restoration transformer with guided deformable attention},
  author={Liang, Jingyun and Fan, Yuchen and Xiang, Xiaoyu and Ranjan, Rakesh and Ilg, Eddy and Green, Simon and Cao, Jiezhang and Zhang, Kai and Timofte, Radu and Gool, Luc V},
  journal={Advances in Neural Information Processing Systems},
  volume={35},
  pages={378--393},
  year={2022}
}

@inproceedings{zhang2024bsstnet,
  title={Blur-aware spatio-temporal sparse transformer for video deblurring},
  author={Zhang, Huicong and Xie, Haozhe and Yao, Hongxun},
  booktitle={Proceedings of the IEEE/CVF Conference on Computer Vision and Pattern Recognition},
  pages={2673--2681},
  year={2024}
}

@inproceedings{rim2020real,
  title={Real-world blur dataset for learning and benchmarking deblurring algorithms},
  author={Rim, Jaesung and Lee, Haeyun and Won, Jucheol and Cho, Sunghyun},
  booktitle={European conference on computer vision},
  pages={184--201},
  year={2020},
  organization={Springer}
}

@inproceedings{rim2022realistic,
  title={Realistic blur synthesis for learning image deblurring},
  author={Rim, Jaesung and Kim, Geonung and Kim, Jungeon and Lee, Junyong and Lee, Seungyong and Cho, Sunghyun},
  booktitle={European conference on computer vision},
  pages={487--503},
  year={2022},
  organization={Springer}
}

@article{zhong2023real,
  title={Real-world video deblurring: A benchmark dataset and an efficient recurrent neural network},
  author={Zhong, Zhihang and Gao, Ye and Zheng, Yinqiang and Zheng, Bo and Sato, Imari},
  journal={International Journal of Computer Vision},
  volume={131},
  number={1},
  pages={284--301},
  year={2023},
  publisher={Springer}
}

@inproceedings{zhu2022deep,
  title={Deep recurrent neural network with multi-scale bi-directional propagation for video deblurring},
  author={Zhu, Chao and Dong, Hang and Pan, Jinshan and Liang, Boyang and Huang, Yuhao and Fu, Lean and Wang, Fei},
  booktitle={Proceedings of the AAAI conference on artificial intelligence},
  volume={36},
  number={3},
  pages={3598--3607},
  year={2022}
}

@inproceedings{kim2024frequency,
  title={Frequency-aware event-based video deblurring for real-world motion blur},
  author={Kim, Taewoo and Cho, Hoonhee and Yoon, Kuk-Jin},
  booktitle={Proceedings of the IEEE/CVF Conference on Computer Vision and Pattern Recognition},
  pages={24966--24976},
  year={2024}
}

@article{lee2024gs,
  title={GS-blur: A 3d scene-based dataset for realistic image deblurring},
  author={Lee, Dongwoo and Park, JoonKyu and Lee, Kyoung M},
  journal={Advances in Neural Information Processing Systems},
  volume={37},
  pages={125394--125415},
  year={2024}
}

@inproceedings{zhang2020rwbi,
  title={Deblurring by realistic blurring},
  author={Zhang, Kaihao and Luo, Wenhan and Zhong, Yiran and Ma, Lin and Stenger, Bjorn and Liu, Wei and Li, Hongdong},
  booktitle={Proceedings of the IEEE/CVF conference on computer vision and pattern recognition},
  pages={2737--2746},
  year={2020}
}

@inproceedings{nah2017gopro,
  title={Deep multi-scale convolutional neural network for dynamic scene deblurring},
  author={Nah, Seungjun and Hyun Kim, Tae and Mu Lee, Kyoung},
  booktitle={Proceedings of the IEEE conference on computer vision and pattern recognition},
  pages={3883--3891},
  year={2017}
}

@InProceedings{nah2019reds,
  author = {Nah, Seungjun and Baik, Sungyong and Hong, Seokil and Moon, Gyeongsik and Son, Sanghyun and Timofte, Radu and Lee, Kyoung Mu},
  title = {NTIRE 2019 Challenge on Video Deblurring and Super-Resolution: Dataset and Study},
  booktitle = {CVPR Workshops},
  month = {June},
  year = {2019}
}

@article{noki2025slomo,
  title={Deblurring in the Wild: A Real-World Dataset from Smartphone High-Speed Videos.},
  author={Noki, Mahdi Mohd Hossain and Mahmud, Syed Mumtahin and Majumder, Prothito Shovon and Al Radi, Abdul Mohaimen and Ali, Md Haider and Khan, Md Mosaddek},
  journal={arXiv preprint arXiv:2506.19445},
  year={2025}
}

@inproceedings{rao2024rethinking,
  title={Rethinking video deblurring with wavelet-aware dynamic transformer and diffusion model},
  author={Rao, Chen and Li, Guangyuan and Lan, Zehua and Sun, Jiakai and Luan, Junsheng and Xing, Wei and Zhao, Lei and Lin, Huaizhong and Dong, Jianfeng and Zhang, Dalong},
  booktitle={European Conference on Computer Vision},
  pages={421--437},
  year={2024},
  organization={Springer}
}

@article{long2024divd,
  title={DIVD: Deblurring with Improved Video Diffusion Model},
  author={Long, Haoyang and Wang, Yan and Wang, Wendong},
  journal={arXiv preprint arXiv:2412.00773},
  year={2024}
}

@inproceedings{lin2024diffbir,
  title={Diffbir: Toward blind image restoration with generative diffusion prior},
  author={Lin, Xinqi and He, Jingwen and Chen, Ziyan and Lyu, Zhaoyang and Dai, Bo and Yu, Fanghua and Qiao, Yu and Ouyang, Wanli and Dong, Chao},
  booktitle={European conference on computer vision},
  pages={430--448},
  year={2024},
  organization={Springer}
}

@inproceedings{kong2025deblurdiff,
  title={DeblurDiff: Real-Word Image Deblurring with Generative Diffusion Models},
  author={Kong, Lingshun and Zou, Dongqing and Wang, Fu Lee and Ren, Jimmy and Wu, Xiaohe and Dong, Jiangxin and Pan, Jinshan and others},
  booktitle={The Thirty-ninth Annual Conference on Neural Information Processing Systems},
  year={2025}
}

@inproceedings{yang2024noise,
  title={Noise Calibration: Plug-and-Play Content-Preserving Video Enhancement Using Pre-trained Video Diffusion Models},
  author={Yang, Qinyu and Chen, Haoxin and Zhang, Yong and Xia, Menghan and Cun, Xiaodong and Su, Zhixun and Shan, Ying},
  booktitle={European Conference on Computer Vision},
  pages={307--326},
  year={2024},
  organization={Springer}
}

@article{yeh2024diffir2vr,
  title={Diffir2vr-zero: Zero-shot video restoration with diffusion-based image restoration models},
  author={Yeh, Chang-Han and Shiu, Hau-Shiang and Lin, Chin-Yang and Wang, Zhixiang and Hsiao, Chi-Wei and Chen, Ting-Hsuan and Liu, Yu-Lun},
  journal={arXiv preprint arXiv:2407.01519},
  year={2024}
}

@inproceedings{li2025tdm,
  title={TDM: Temporally-Consistent Diffusion Model for All-in-One Real-World Video Restoration},
  author={Li, Yizhou and Liu, Zihua and Monno, Yusuke and Okutomi, Masatoshi},
  booktitle={International Conference on Multimedia Modeling},
  pages={155--169},
  year={2025},
  organization={Springer}
}

@inproceedings{peebles2023dit,
  title={Scalable diffusion models with transformers},
  author={Peebles, William and Xie, Saining},
  booktitle={Proceedings of the IEEE/CVF international conference on computer vision},
  pages={4195--4205},
  year={2023}
}

@article{wan2025wan,
  title={Wan: Open and advanced large-scale video generative models},
  author={Wan, Team and Wang, Ang and Ai, Baole and Wen, Bin and Mao, Chaojie and Xie, Chen-Wei and Chen, Di and Yu, Feiwu and Zhao, Haiming and Yang, Jianxiao and others},
  journal={arXiv preprint arXiv:2503.20314},
  year={2025}
}

@article{salimans2022progressive,
  title={Progressive distillation for fast sampling of diffusion models},
  author={Salimans, Tim and Ho, Jonathan},
  journal={arXiv preprint arXiv:2202.00512},
  year={2022}
}

@inproceedings{yin2024dmd,
  title={One-step diffusion with distribution matching distillation},
  author={Yin, Tianwei and Gharbi, Micha{\"e}l and Zhang, Richard and Shechtman, Eli and Durand, Fredo and Freeman, William T and Park, Taesung},
  booktitle={Proceedings of the IEEE/CVF conference on computer vision and pattern recognition},
  pages={6613--6623},
  year={2024}
}

@article{hu2022lora,
  title={Lora: Low-rank adaptation of large language models.},
  author={Hu, Edward J and Shen, Yelong and Wallis, Phillip and Allen-Zhu, Zeyuan and Li, Yuanzhi and Wang, Shean and Wang, Liang and Chen, Weizhu and others},
  journal={Iclr},
  volume={1},
  number={2},
  pages={3},
  year={2022}
}

@article{su2024roformer,
  title={Roformer: Enhanced transformer with rotary position embedding},
  author={Su, Jianlin and Ahmed, Murtadha and Lu, Yu and Pan, Shengfeng and Bo, Wen and Liu, Yunfeng},
  journal={Neurocomputing},
  volume={568},
  pages={127063},
  year={2024},
  publisher={Elsevier}
}

@article{wei2025videorope,
  title={Videorope: What makes for good video rotary position embedding?},
  author={Wei, Xilin and Liu, Xiaoran and Zang, Yuhang and Dong, Xiaoyi and Zhang, Pan and Cao, Yuhang and Tong, Jian and Duan, Haodong and Guo, Qipeng and Wang, Jiaqi and others},
  journal={arXiv preprint arXiv:2502.05173},
  year={2025}
}

@article{lipman2022flow,
  title={Flow matching for generative modeling},
  author={Lipman, Yaron and Chen, Ricky TQ and Ben-Hamu, Heli and Nickel, Maximilian and Le, Matt},
  journal={arXiv preprint arXiv:2210.02747},
  year={2022}
}

@inproceedings{yeshwanth2023scannet++,
  title={Scannet++: A high-fidelity dataset of 3d indoor scenes},
  author={Yeshwanth, Chandan and Liu, Yueh-Cheng and Nie{\ss}ner, Matthias and Dai, Angela},
  booktitle={Proceedings of the IEEE/CVF International Conference on Computer Vision},
  pages={12--22},
  year={2023}
}

@misc{InteriorGS2025,
  title        = {InteriorGS: A 3D Gaussian Splatting Dataset of Semantically Labeled Indoor Scenes},
  author       = {SpatialVerse Research Team, Manycore Tech Inc.},
  year         = {2025},
  howpublished = {\url{https://huggingface.co/datasets/spatialverse/InteriorGS}}
}

@inproceedings{peng2024bags,
  title={Bags: Blur agnostic gaussian splatting through multi-scale kernel modeling},
  author={Peng, Cheng and Tang, Yutao and Zhou, Yifan and Wang, Nengyu and Liu, Xijun and Li, Deming and Chellappa, Rama},
  booktitle={European Conference on Computer Vision},
  pages={293--310},
  year={2024},
  organization={Springer}
}

@inproceedings{lee2024deblurring3dgs,
  title={Deblurring 3d gaussian splatting},
  author={Lee, Byeonghyeon and Lee, Howoong and Sun, Xiangyu and Ali, Usman and Park, Eunbyung},
  booktitle={European Conference on Computer Vision},
  pages={127--143},
  year={2024},
  organization={Springer}
}

@article{kerbl20233dgs,
  title={3d gaussian splatting for real-time radiance field rendering.},
  author={Kerbl, Bernhard and Kopanas, Georgios and Leimk{\"u}hler, Thomas and Drettakis, George and others},
  journal={ACM Trans. Graph.},
  volume={42},
  number={4},
  pages={139--1},
  year={2023}
}

@inproceedings{su2017dvd,
  title={Deep video deblurring for hand-held cameras},
  author={Su, Shuochen and Delbracio, Mauricio and Wang, Jue and Sapiro, Guillermo and Heidrich, Wolfgang and Wang, Oliver},
  booktitle={Proceedings of the IEEE conference on computer vision and pattern recognition},
  pages={1279--1288},
  year={2017}
}

@article{chu2020learning,
  title={Learning temporal coherence via self-supervision for GAN-based video generation},
  author={Chu, Mengyu and Xie, You and Mayer, Jonas and Leal-Taix{\'e}, Laura and Thuerey, Nils},
  journal={ACM Transactions on Graphics (TOG)},
  volume={39},
  number={4},
  pages={75--1},
  year={2020},
  publisher={ACM New York, NY, USA}
}

@inproceedings{zhang2018lpips,
  title={The unreasonable effectiveness of deep features as a perceptual metric},
  author={Zhang, Richard and Isola, Phillip and Efros, Alexei A and Shechtman, Eli and Wang, Oliver},
  booktitle={Proceedings of the IEEE conference on computer vision and pattern recognition},
  pages={586--595},
  year={2018}
}

@article{heusel2017fid,
  title={Gans trained by a two time-scale update rule converge to a local nash equilibrium},
  author={Heusel, Martin and Ramsauer, Hubert and Unterthiner, Thomas and Nessler, Bernhard and Hochreiter, Sepp},
  journal={Advances in neural information processing systems},
  volume={30},
  year={2017}
}

@inproceedings{ke2021musiq,
  title={Musiq: Multi-scale image quality transformer},
  author={Ke, Junjie and Wang, Qifei and Wang, Yilin and Milanfar, Peyman and Yang, Feng},
  booktitle={Proceedings of the IEEE/CVF international conference on computer vision},
  pages={5148--5157},
  year={2021}
}

@article{mittal2012niqe,
  title={Making a “completely blind” image quality analyzer},
  author={Mittal, Anish and Soundararajan, Rajiv and Bovik, Alan C},
  journal={IEEE Signal processing letters},
  volume={20},
  number={3},
  pages={209--212},
  year={2012},
  publisher={IEEE}
}

@inproceedings{li2023shiftnet,
  title={A simple baseline for video restoration with grouped spatial-temporal shift},
  author={Li, Dasong and Shi, Xiaoyu and Zhang, Yi and Cheung, Ka Chun and See, Simon and Wang, Xiaogang and Qin, Hongwei and Li, Hongsheng},
  booktitle={Proceedings of the IEEE/CVF Conference on Computer Vision and Pattern Recognition},
  pages={9822--9832},
  year={2023}
}

@article{liang2024vrt,
  title={Vrt: A video restoration transformer},
  author={Liang, Jingyun and Cao, Jiezhang and Fan, Yuchen and Zhang, Kai and Ranjan, Rakesh and Li, Yawei and Timofte, Radu and Van Gool, Luc},
  journal={IEEE Transactions on Image Processing},
  volume={33},
  pages={2171--2182},
  year={2024},
  publisher={IEEE}
}

@inproceedings{ma2022deblur,
  title={Deblur-nerf: Neural radiance fields from blurry images},
  author={Ma, Li and Li, Xiaoyu and Liao, Jing and Zhang, Qi and Wang, Xuan and Wang, Jue and Sander, Pedro V},
  booktitle={Proceedings of the IEEE/CVF conference on computer vision and pattern recognition},
  pages={12861--12870},
  year={2022}
}

@article{ye2025gsplat,
  title={gsplat: An open-source library for Gaussian splatting},
  author={Ye, Vickie and Li, Ruilong and Kerr, Justin and Turkulainen, Matias and Yi, Brent and Pan, Zhuoyang and Seiskari, Otto and Ye, Jianbo and Hu, Jeffrey and Tancik, Matthew and others},
  journal={Journal of Machine Learning Research},
  volume={26},
  number={34},
  pages={1--17},
  year={2025}
}

@article{kheradmand20243d,
  title={3d gaussian splatting as markov chain monte carlo},
  author={Kheradmand, Shakiba and Rebain, Daniel and Sharma, Gopal and Sun, Weiwei and Tseng, Yang-Che and Isack, Hossam and Kar, Abhishek and Tagliasacchi, Andrea and Yi, Kwang Moo},
  journal={Advances in Neural Information Processing Systems},
  volume={37},
  pages={80965--80986},
  year={2024}
}

@article{lin2025depth,
  title={Depth anything 3: Recovering the visual space from any views},
  author={Lin, Haotong and Chen, Sili and Liew, Junhao and Chen, Donny Y and Li, Zhenyu and Shi, Guang and Feng, Jiashi and Kang, Bingyi},
  journal={arXiv preprint arXiv:2511.10647},
  year={2025}
}

@article{chen2026anyrecon,
  title={AnyRecon: Arbitrary-View 3D Reconstruction with Video Diffusion Model},
  author={Chen, Yutian and Guo, Shi and Jin, Renbiao and Yang, Tianshuo
          and Cai, Xin and Luo, Yawen and Yang, Mingxin and Yu, Mulin
          and Xu, Linning and Xue, Tianfan},
  journal={arXiv preprint arXiv:2604.19747},
  year={2026}
}

\clearpage

\renewcommand{\thefigure}{S\arabic{figure}}
\renewcommand{\thetable}{S\arabic{table}}
\renewcommand{\theequation}{S\arabic{equation}}
\renewcommand{\thesection}{S\arabic{section}}
\renewcommand{\thesubsection}{S\arabic{section}.\arabic{subsection}}
\setcounter{figure}{0}
\setcounter{table}{0}
\setcounter{equation}{0}
\setcounter{section}{0}

\title{RealVDeblur: Supplementary Material}
\author{}
\institute{}
\maketitle
\begin{sloppypar}

\section{Implementation Details}
\label{sec:supp_impl}
\noindent\textbf{Frame-wise VAE Encoding.}
Wan-VAE employs a causal temporal compression architecture that downsamples the input sequence by a factor of $4$ along the temporal dimension, mapping $(1{+}T)$ input frames to $(1{+}T/4)$ latents. 
Notably, the first frame is encoded independently without temporal downsampling. 
When the input reduces to a single image ($T{=}0$), the VAE naturally reduces to a purely 2D encoder. 
Leveraging this property, we treat each video frame as an independent image and encode them separately, yielding $T$-frame-in, $T$-frame-out latents that avoid cross-frame entanglement introduced by temporal compression.

\noindent\textbf{Condition Injection.}
Given blurry input frames, a frozen frame-wise VAE encoder first produces the blur latent $z_{\mathrm{blur}}$.
We then apply a lightweight 3D convolutional patch embedding layer (16$\rightarrow$1536 channels, kernel size $1\times2\times2$, stride $1\times2\times2$) to transform the blur latent into a conditioning tensor whose spatial resolution matches the patchified noise embedding used by the DiT.
The resulting conditioning tensor is directly added to the noise embedding before being processed by the transformer blocks.

\noindent\textbf{Training Configuration.}
We fine-tune the LoRA parameters (rank 64) and the condition injection network on OmniBlur for 5,000 iterations using $32\times$ A100 GPUs with a total batch size of 32.
We adopt a mixed-resolution training strategy that combines different sequence lengths and spatial resolutions within each batch. Specifically, we use 10-frame sequences at 1280$\times$720 and 20-frame sequences at 960$\times$544.
For one-step distillation, both the Student and Critic LoRA are initialized from the Teacher LoRA while keeping the DiT backbone frozen. Due to GPU memory constraints, the VAE decoder is applied only to a random subset of 5 frames per batch when computing pixel-space supervision.

\section{OmniBlur Dataset Details}
\label{sec:supp_dataset}
\subsection{Dataset Statistics}
Our OmniBlur training dataset is constructed from two complementary sources: 3DGS-rendered scenes and high-frame-rate videos.

For the 3DGS-rendered data, we use 956 scenes from ScanNet++~\cite{yeshwanth2023scannet++} and 1,000 scenes from InteriorGS~\cite{interiorgs2025}. These scenes cover camera-motion blur, defocus blur, and their compound combinations. From them, we render and sample approximately 53,000 training clips (20 frames each). Among the 3DGS-rendered data, camera-motion blur accounts for ${\sim}$40\%, defocus blur for ${\sim}$40\%, and compound blur for ${\sim}$20\%.
For the high-frame-rate video data, we sample 3,000 short videos (20 frames each) from 20 GoPro~\cite{nah2017gopro}, 240 REDS~\cite{nah2019reds}, and 734 SloMoDeblur~\cite{noki2025slomo} scenes to cover object-motion blur.
Overall, OmniBlur contains ${\sim}$2,000 3DGS scenes and ${\sim}$3,000 high-frame-rate videos, producing about 56,000 training clips with approximately 1.1M total frames.

\begin{figure}[htbp]
  \centering
  \includegraphics[width=\textwidth]{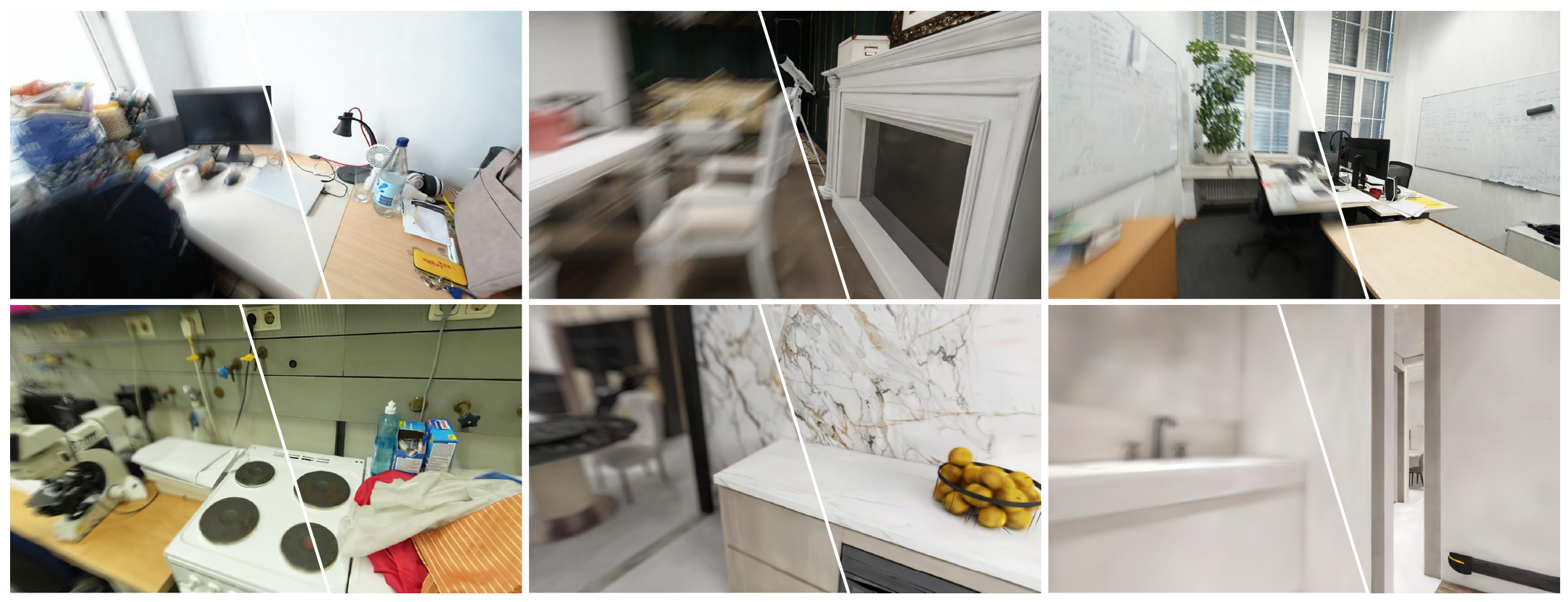}
  \caption{Examples of synthesized blur from the OmniBlur dataset. Top: examples of camera-motion blur. Bottom: examples of defocus blur generated with a thin-lens model. In each example, the left side of the diagonal line shows the sharp rendering, while the right side shows the synthesized blur.}
  \label{fig:supp_synthesis}
\end{figure}

\subsection{Blur Synthesis Parameters}
The overall synthesis pipeline is described in Sec.~3.3 of the main paper.
Here we provide the specific parameter settings.

\noindent\textbf{Camera-Motion Blur.}
For each frame in the generated camera trajectory, we introduce smooth 6-DoF perturbations using B\'{e}zier curves with randomly sampled degrees (1--4) to simulate hand-held camera shake.
The motion magnitude is adapted to each scene by scaling the average inter-frame camera displacement from the original trajectory.
Within each exposure interval, $K{=}121$ sub-frames are rendered along the perturbed trajectory and averaged in linear RGB space to produce the blurry frame.
The corresponding sharp frame is defined as the center sub-frame $I_{\mathrm{sub}}^{(K/2)}$.

\noindent\textbf{Defocus Blur.}
We follow a thin-lens camera model. For each sequence, a focal point is randomly sampled and its depth is obtained from the 3DGS-rendered depth map. Multiple viewpoints ($N{=}32$) are sampled on the circular aperture disk around the center camera and redirected toward the same focal point.
The rendered views are averaged in linear RGB space to produce depth-dependent defocus blur.

\noindent\textbf{Compound Blur.}
Both camera-motion and defocus degradations are applied simultaneously.
To balance rendering cost, the camera-motion sub-frame count is reduced to $K{=}61$ and the defocus sample count is reduced to $N{=}16$ for compound blur, resulting in $K{\times}N{=}61{\times}16{=}976$ sub-renderings per frame.

\noindent\textbf{ISP-Aware Augmentation.}
Rendered images from 3DGS are noise-free and bypass the camera image signal processing (ISP), creating a domain gap with real camera captures. 
Following RSBlur~\cite{rim2022realistic}, we simulate a simplified RAW imaging pipeline. 
Specifically, rendered sRGB images are first converted to the linear domain and mapped to camera RGB space. 
The images are then mosaiced into a Bayer pattern with a randomly selected layout. 
Inverse white balance is applied to simulate scene illumination, after which Poisson--Gaussian noise is injected in the RAW domain with randomly scaled noise levels to emulate different sensor ISO settings. 
Finally, a forward ISP (white balance, demosaicing, color correction, and sRGB encoding) is applied to produce realistic training images.

\begin{figure}[htbp]
  \centering
  \includegraphics[width=\textwidth]{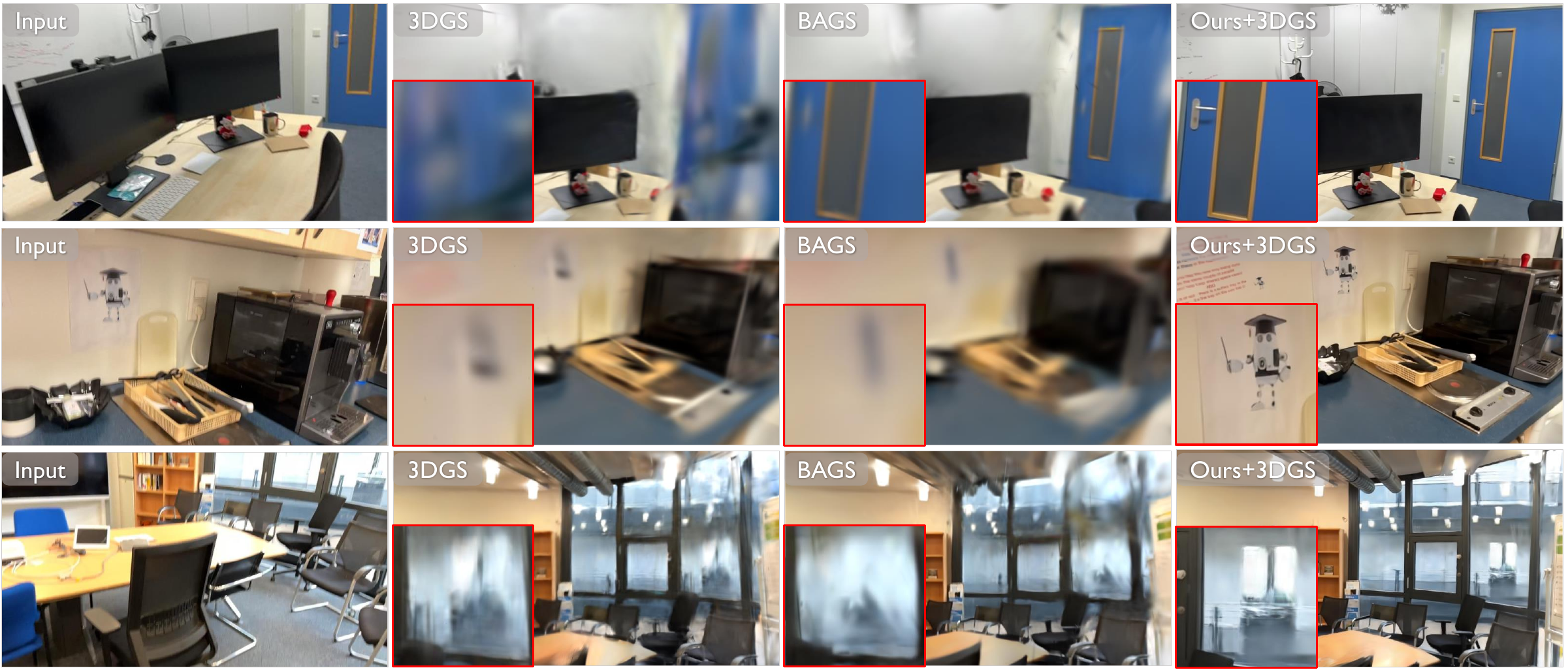}
  \caption{Visual comparison of 3DGS reconstruction on large-scale real-world scenes. Directly training on blurry inputs or using joint optimization (BAGS) produces severe floaters and oversmoothed textures. Preprocessing with RealVDeblur (Ours) yields significantly sharper reconstructions. Red insets highlight representative regions.}
  \label{fig:supp_largescene}
\end{figure}

\section{Downstream Task Evaluation}
\label{sec:supp_downstream}
\subsection{3DGS Reconstruction Setup}
We evaluate on the DeblurNeRF~\cite{ma2022deblur} dataset, which contains 10 scenes with real camera-motion blur and 10 scenes with real defocus blur. All experiments are conducted at a resolution of $1200\times800$.
For cascaded pipelines (VRT~\cite{liang2022vrt} + 3DGS, RealVDeblur + 3DGS), we first treat the multi-view training images as consecutive frames and apply the deblurring model to all views.  We then run COLMAP on the deblurred images to estimate camera poses and initialize the point cloud. 
A 3DGS model is then trained using the \texttt{gsplat}~\cite{ye2025gsplat} implementation with the MCMC~\cite{kheradmand20243d} densification strategy and default hyperparameters for 20,000 iterations. 
For end-to-end baselines, we use the official implementations of BAGS~\cite{peng2024bags} and Deblurring-3DGS~\cite{lee2024deblurring3dgs} with their default settings. The camera poses and initial point clouds provided by the Deblur-NeRF dataset are used for initialization, which are reconstructed from the original blurry images.

\subsection{Robustness on Larger-Scale Real Scenes}
To further evaluate the practical robustness of RealVDeblur, we perform 3DGS reconstruction on larger-scale real-world scenes without ground truth. Specifically, we select three scenes from the ScanNet++ iPhone test set~\cite{yeshwanth2023scannet++} and extract partially blurry video sequences captured using handheld mobile phones. Unlike the Deblur-NeRF dataset, which contains relatively small forward-facing camera motions, these sequences cover room-scale environments with more complex camera trajectories and natural motion blur.

We compare our method with 3DGS, and BAGS under the same setup. 
As shown in Fig.~\ref{fig:supp_largescene}, directly training 3DGS on blurry inputs produces significant floater artifacts and blurred textures. 
After RealVDeblur preprocessing, the reconstructed scenes exhibit substantially sharper textures and greatly reduced floater artifacts, demonstrating strong robustness in realistic capture scenarios.
In contrast, joint optimization methods are particularly sensitive to blur, as inaccurate pose estimation derived from blurred inputs often leads to suboptimal initialization. Moreover, apart from initialization difficulties, these approaches inherently struggle to scale to large-scale scenes due to their limited capacity to handle the increased optimization complexity of room-scale environments

\subsection{Depth Estimation}
\begin{figure}[htbp]
  \centering
  \includegraphics[width=\textwidth]{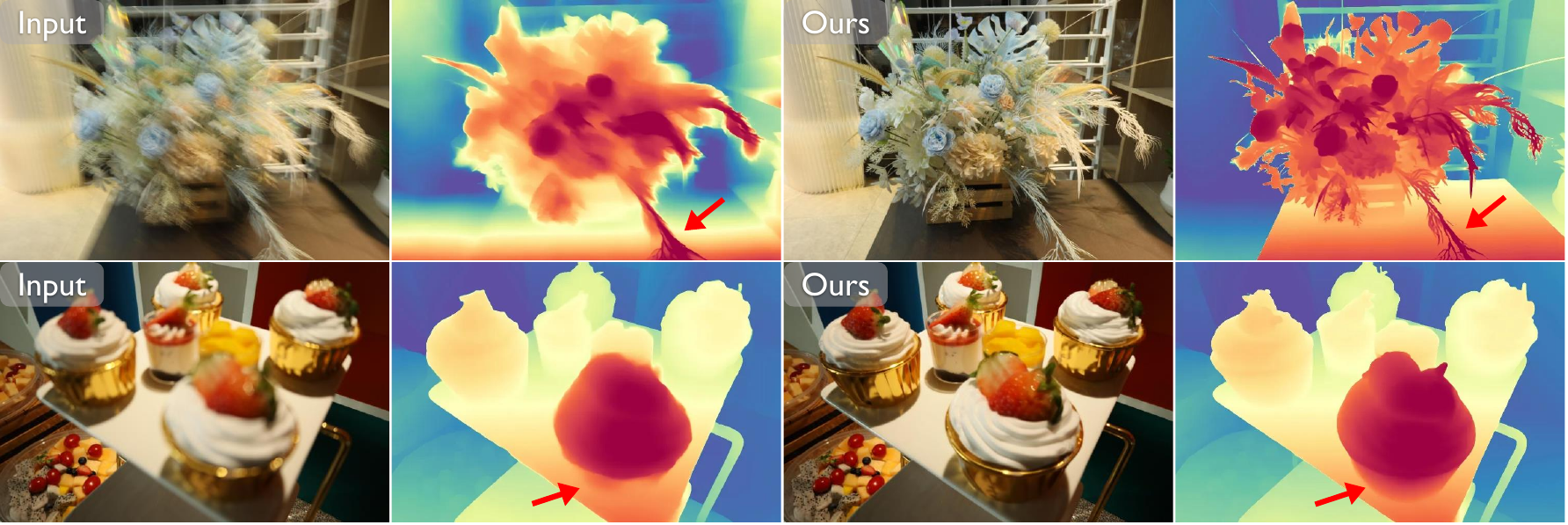}
  \caption{Impact of video deblurring on downstream depth estimation. Depth maps estimated from raw blurry inputs (second column) exhibit distorted boundaries and inconsistent geometry. By using RealVDeblur(Ours) as a preprocessing step, the depth maps(fourth column) exhibit significantly sharper edges and improved structural integrity. Red arrows highlight regions where our method effectively restores thin structures and clear depth discontinuities that are otherwise lost in blur.}
  \label{fig:supp_depth}
\end{figure}

Motion blur severely corrupts edge and structural information, directly impairing downstream depth estimation.
We evaluate the impact of video deblurring as a preprocessing step for depth estimation using DepthAnythingV3~\cite{lin2025depth}.
As shown in Fig.~\ref{fig:supp_depth}, depth maps estimated from blurry inputs suffer from blurred depth boundaries and inaccurate geometry.
After RealVDeblur deblurring, the estimated depth maps exhibit significantly sharper edges, improved structural completeness, and more consistent depth in planar regions, demonstrating that high-quality deblurring is essential for reliable downstream 3D perception.
% ====================================================================

\begin{figure}[htbp]
  \centering
  \includegraphics[width=\textwidth]{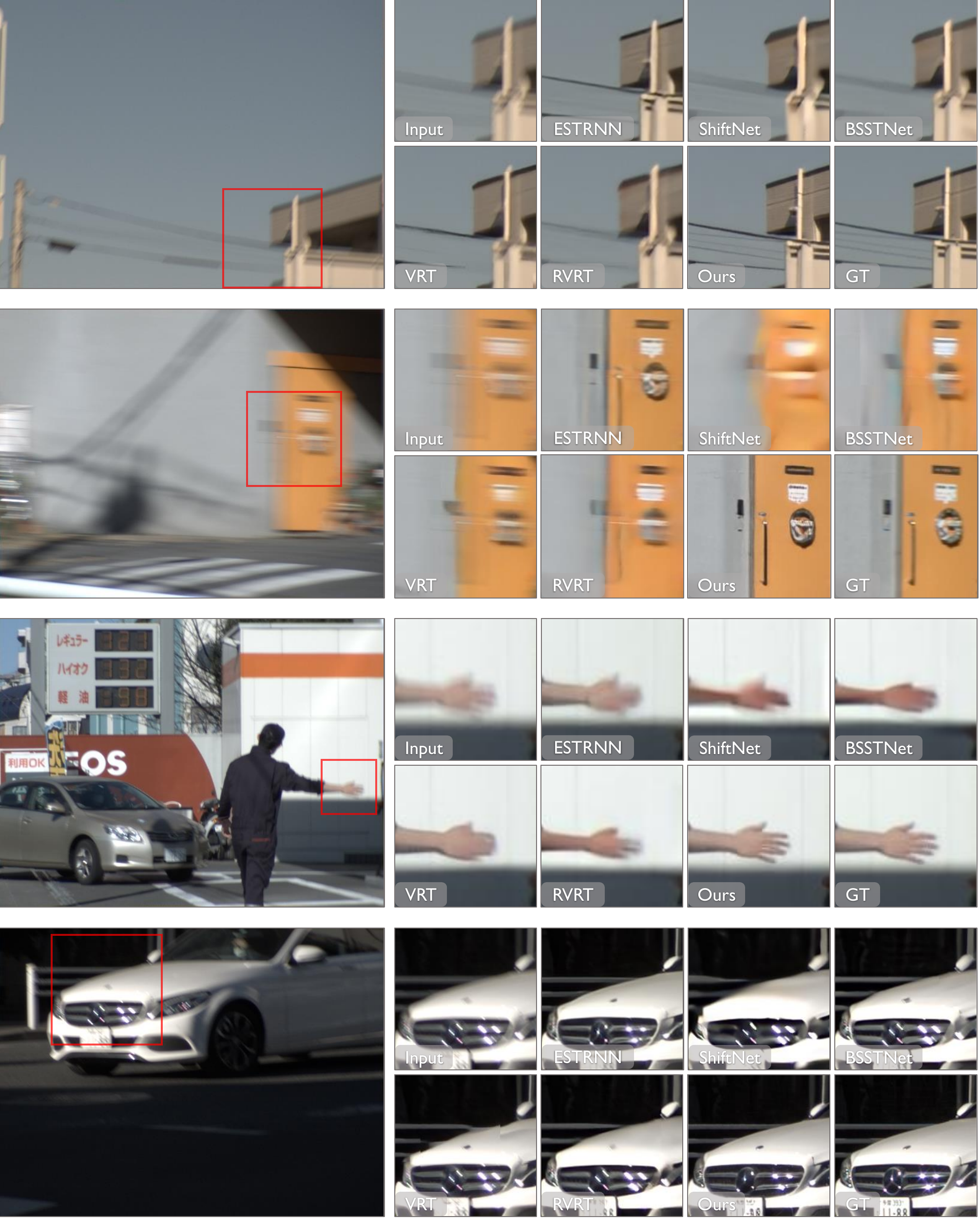}
  \caption{Additional visual comparisons on the BSD benchmark.}
  \label{fig:supp_bsd}
\end{figure}
\begin{figure}[htbp]
  \centering
  \includegraphics[width=\textwidth]{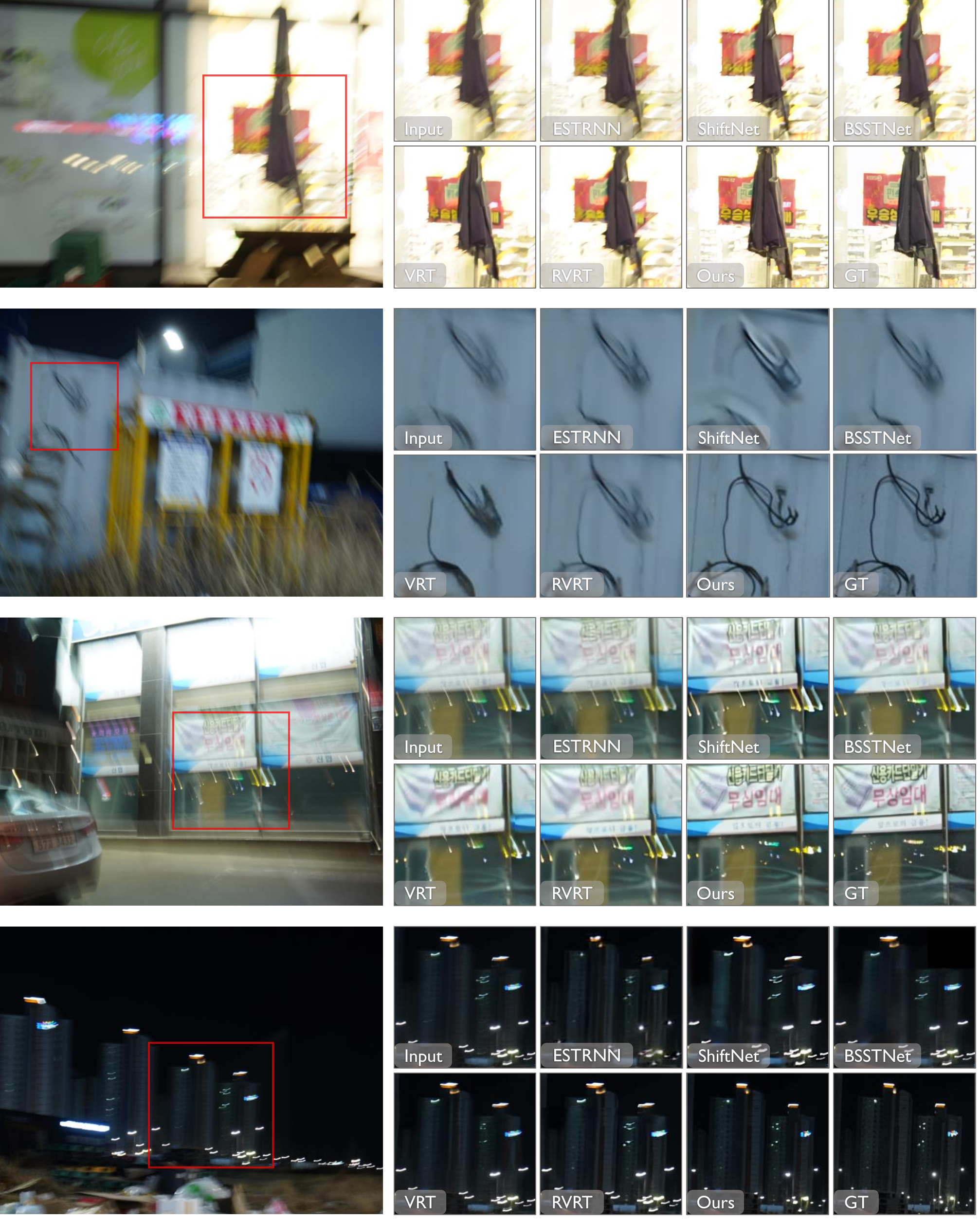}
  \caption{Additional visual comparisons on the RealBlur benchmark.}
  \label{fig:supp_realblur}
\end{figure}
\begin{figure}[htbp]
  \centering
  \includegraphics[width=\textwidth]{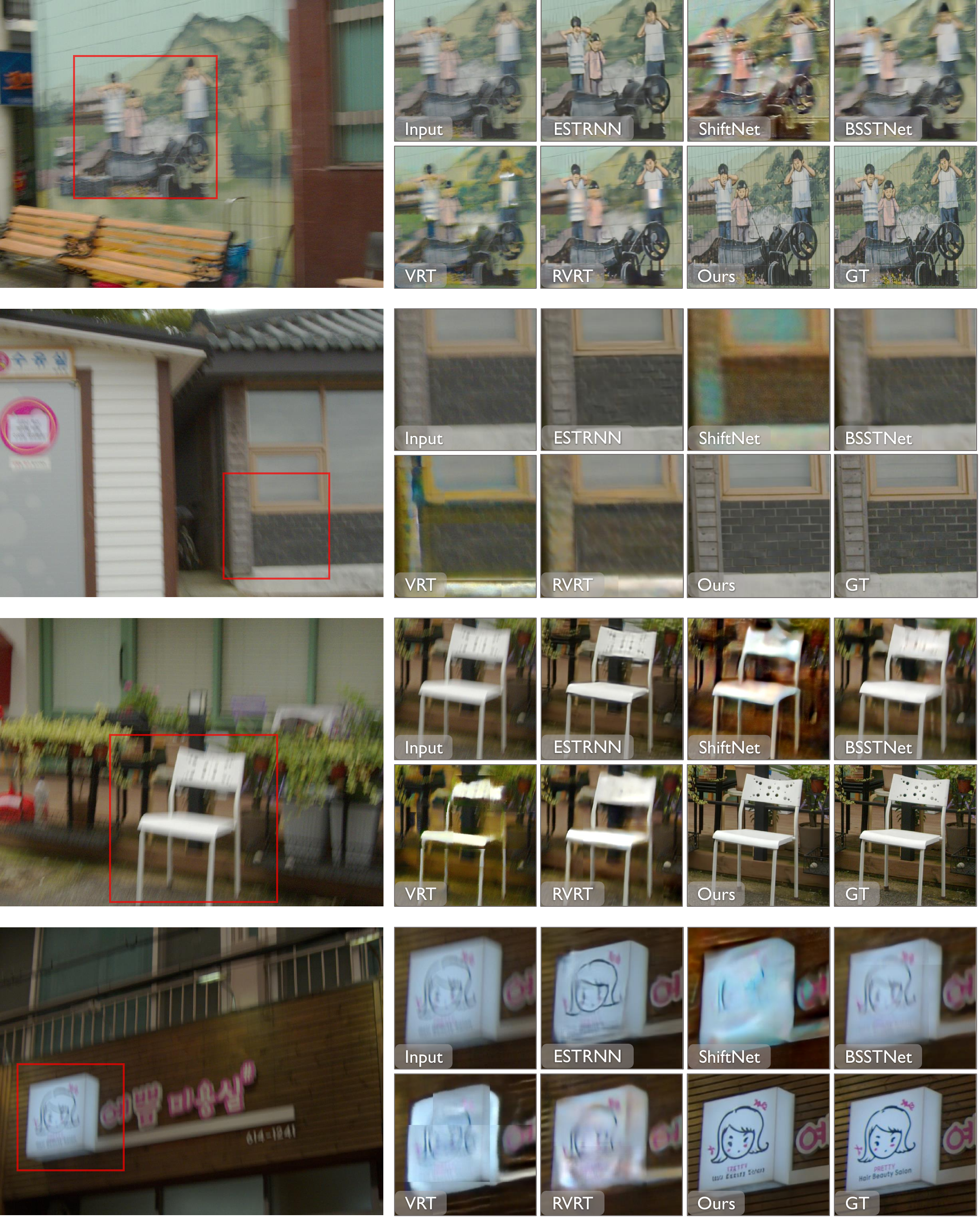}
  \caption{Additional visual comparisons on the RSBlur benchmark.}
  \label{fig:supp_rsblur}
\end{figure}
\begin{figure}[htbp]
  \centering
  \includegraphics[width=\textwidth]{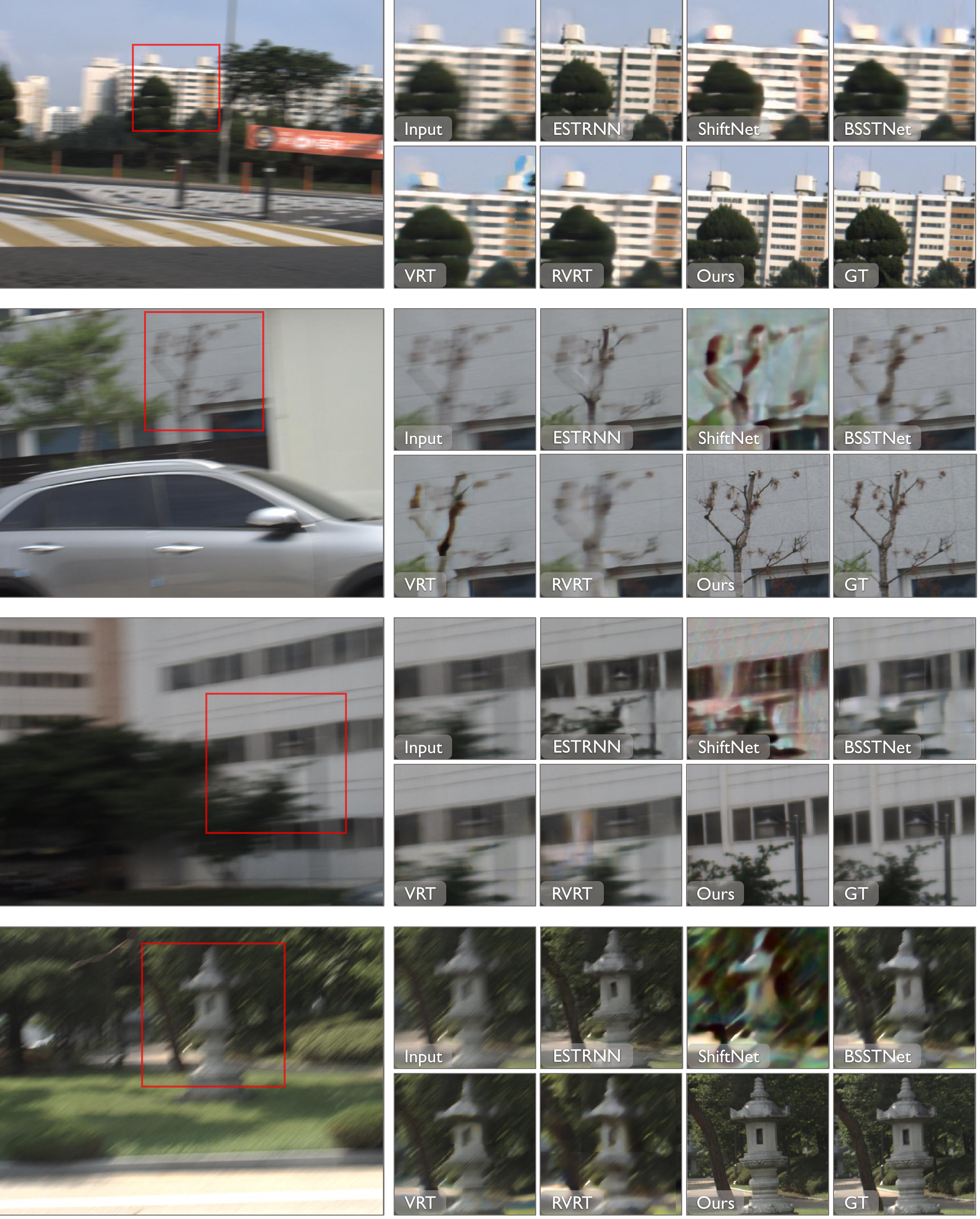}
  \caption{Additional visual comparisons on the FEVD benchmark.}
  \label{fig:supp_fevd}
\end{figure}

\section{Additional Qualitative Results}
\label{sec:supp_qualitative}
% \subsection{Visual Comparisons on Each Benchmark}
We provide additional qualitative comparisons on each evaluation benchmark in Figs.~\ref{fig:supp_bsd}--\ref{fig:supp_fevd}.
For each example, we show the blurry input alongside representative baselines and our RealVDeblur result, with zoomed-in crops highlighting fine-detail restoration.

\section{Failure Cases and Limitations}
\begin{figure}[htbp]
  \centering
  \includegraphics[width=\textwidth]{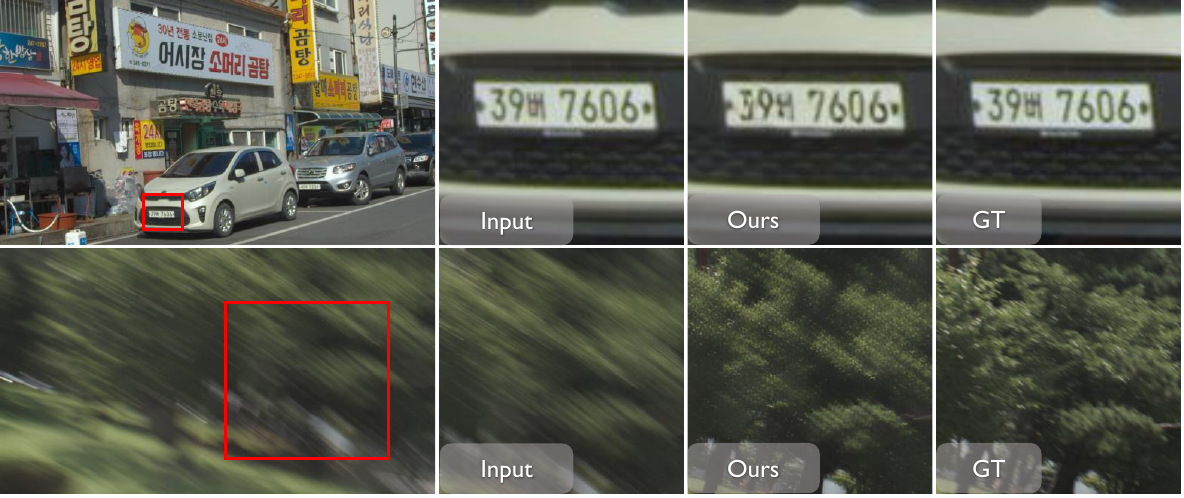}
  \caption{Representative failure cases. (a)~Structural distortion in small text restoration under moderate blur. (b)~Limited restoration of fine structural details in high-frequency regions under extreme motion blur.}
  \label{fig:supp_failure}
\end{figure}

We present representative failure cases to illustrate the limitations of our method.
\noindent\textbf{Structural sensitivity in text restoration.} While RealVDeblur excels at restoring natural textures, it faces challenges with small, structured patterns like text. As shown in the top row of Fig.~\ref{fig:supp_failure}, even when the motion blur is not the dominant degradation, the reconstructed characters may still exhibit structural distortions. This highlights a known trade-off in latent-based generative models: while the latent space provides a powerful prior for natural scene synthesis, it may lack the extreme, pixel-level geometric precision required to satisfy the rigid constraints of typography. 

\noindent\textbf{Limited restoration of high-frequency details under extreme blur.} When motion blur is excessive, the model fails to faithfully restore high-frequency regions, such as the complex foliage shown in the bottom row of Fig.~\ref{fig:supp_failure}. While the generative prior helps produce sharp visual results, the reconstructed textures often lack structural fidelity to the ground truth.

\end{sloppypar}

\end{document}